\documentclass[lettersize,journal]{IEEEtran}
\usepackage{amsmath,amsfonts}
\usepackage{algorithmic}
\usepackage{array}
\usepackage[caption=false,font=footnotesize,labelfont=rm,textfont=rm]{subfig}
\usepackage{textcomp}
\usepackage{stfloats}
\usepackage{url}
\usepackage{verbatim}
\usepackage{graphicx}
\def\BibTeX{{\rm B\kern-.05em{\sc i\kern-.025em b}\kern-.08em
    T\kern-.1667em\lower.7ex\hbox{E}\kern-.125emX}}
\usepackage{balance}

\usepackage{xspace}
\usepackage{upgreek}
\usepackage{adjustbox}
\usepackage{booktabs}
\usepackage{makecell}
\usepackage{hyperref}
\usepackage{mathtools}
\usepackage{orcidlink}
\usepackage{tabularx}
\usepackage{multirow}
\usepackage{pifont}
\newcommand{\ours}{GVCHR\xspace}
\newcommand{\ie}{\textit{i.e.}\xspace}

\newcommand{\wrt}{\textit{w.r.t.}\xspace}

\begin{document}
\title{Generative Video Compression Based on Hierarchical Referencing}
\author{\author{Daowen Li$^{\orcidlink{0000-0001-7805-9967}}$$^{\dagger}$, Ding Ding$^{\orcidlink{0009-0009-3873-1195}}$$^{\dagger}$, Zifu Zhang$^{\orcidlink{0009-0002-6787-5701}}$, \IEEEmembership{Student Member, IEEE}, Kai Li, and Ying Chen$^{\orcidlink{0000-0002-1620-9904}}$, \IEEEmembership{Senior Member, IEEE}
\thanks{Date of current version \today.
}
\thanks{$^{\dagger}$ Equal contribution. (Corresponding author: Ying Chen.)}
\thanks{Daowen Li, Ding Ding, Zifu Zhang, Kai Li, and Ying Chen are with Alibaba Group, Hangzhou 311121, China (e-mail: \{lidaowen.ldw, liangjie.dd, zhangzifu.zzf, kaishi.lk, YingChen\}@alibaba-inc.com).
}
}}

\markboth{Journal of \LaTeX\ Class Files,~Vol.~18, No.~9, September~2020}%
{How to Use the IEEEtran \LaTeX \ Templates}

\maketitle

\begin{abstract}
Diffusion-based generative video compression has emerged as a promising paradigm to improve perceptual quality, where latent frames are required to be encoded efficiently while serving as denoising conditions. However, existing methods neither carefully design reference and quality structures during latent coding nor account for the impact of frame-level quality variation on denoising procedure, which limits coding efficiency and aggravates artifact propagation during generative reconstruction. In this paper, we propose \textbf{\ours}, \textbf{G}enerative \textbf{V}ideo \textbf{C}ompression \textbf{based} on \textbf{Hi}erarchical \textbf{R}eferencing. The key idea is to organize latent frames hierarchically, where the selected high-quality references benefit both latent coding and generative reconstruction. In latent coding, \ours couples a hierarchical reference structure with a hierarchical quality structure, assigning more bits to lower-layer frames that are reused more frequently as references. Built on this design, we introduce Hierarchical Temporal Context Mining to exploits complementary short- and long-term temporal context for effective latent coding. In generative reconstruction, the coding-side hierarchy is incorporated into a Hierarchical Attentive Adapter which is attached to a video diffusion transformer. This adapter uses hierarchical attention to restrict each latent frame to attend only to the same- or lower-layer references, thereby reducing artifact propagation during denoising. Experiments validate \ours on multiple benchmarks. Compared with the previous state-of-the-art method, \ours achieves 50.5\% and 54.0\% BD-rate gains in terms of LPIPS and DISTS, respectively, while also delivering clearly improved visual quality.

\end{abstract}

\begin{IEEEkeywords}
Generative video compression, Diffusion model, Hierarchical reference structure, Perceptual quality
\end{IEEEkeywords}

\section{Introduction}

\begin{figure*}[t]
\centering
\includegraphics[width=\textwidth]{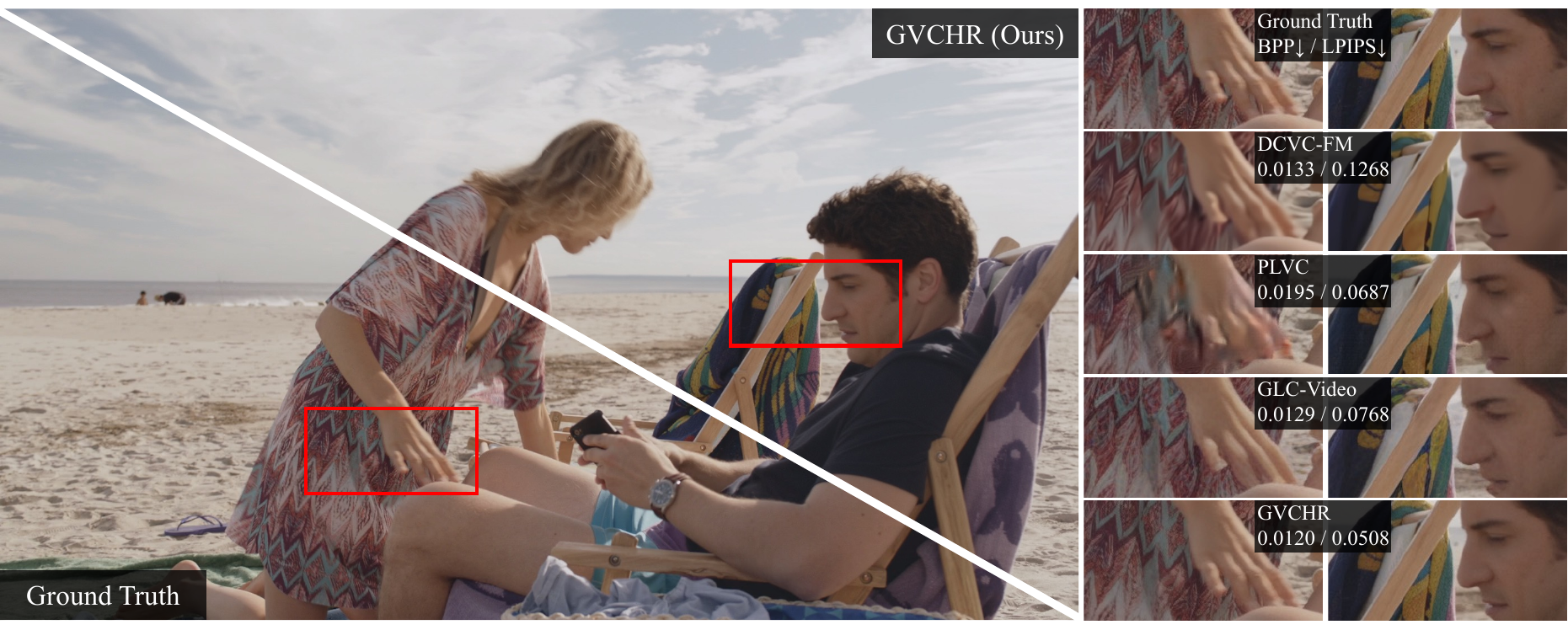}
\vspace{-20pt}
\caption{Visual comparisons with open-source baselines on the \textit{videoSRC24} sequence from MCL-JCV.}
\vspace{-10pt}
\label{fig:videoSRC24}
\end{figure*}

\IEEEPARstart{V}{ideo} traffic dominates today's Internet, highlighting the importance of efficient video compression methods for saving storage and transmission costs. Existing traditional codecs~\cite{avc,hevc,vvc} and most neural video compression (NVC) methods~\cite{dcvc,dcvc-tcm,dcvc-dc,dcvc-fm,dcvc-rt} are primarily optimized for fidelity-oriented objectives, such as PSNR or mean squared error, which often lead to blurry structures and visually unsatisfactory artifacts at low bitrates. This mismatch between distortion minimization and human perception has motivated increasing interest in generative video compression (GVC), where generative priors are exploited to reconstruct more realistic and perceptually pleasing videos. Recent advances in diffusion models~\cite{wan,vace,cogvideox} have further strengthened this direction, owing to their superior ability to synthesize photo-realistic spatial and temporal details. As a result, diffusion-based GVC has emerged as a promising paradigm~\cite{diffvc,gnvc-vd}, in which compressed latent representations are transmitted and then used to guide generative reconstruction at the decoder.

A typical diffusion-based GVC pipeline first encodes a source video into latent representations using a variational autoencoder (VAE)~\cite{vae}, and then compresses them with a learned latent codec. During reconstruction, the decoded latent frames are fed to a diffusion model as conditions for denoising, and the denoised latent frames are finally mapped back to the pixel domain by the VAE decoder. In latent coding, the clean latent frames should be compressed efficiently, which depends on how previously decoded frames are organized and exploited as temporal context, as well as how coding quality is allocated across frames. In particular, frames that play more important reference roles in providing context should be preserved with higher quality. In generative reconstruction, the decoded latent frames become lossy conditions for denoising, where the key issue is to organize their reference dependencies so as to reduce the propagation of coding artifacts across frames. Therefore, how to organize latent frames for both efficient coding and effective generative reconstruction becomes a central issue in diffusion-based GVC.

However, existing diffusion-based GVC methods do not organize latent frames in a way that fully serves both stages~\cite{gnvc-vd,diffvc}. In latent coding, they often rely on simple frame-by-frame referencing, or introduce frame-level quality variation without explicitly coupling it to frame's reference role. As a result, reference frames that are important to enrich temporal context, might not be preserved with higher quality. In generative reconstruction, decoded latent frames are typically injected into image or video diffusion models through preceding-frame conditioning or full attention across frames. Due to these lossy conditioning signals, unrestricted cross-frame interaction can spread artifacts from less reliable frames and further amplify them during denoising. These limitations suggest that hierarchical organization of latent frames may benefit diffusion-based GVC. Hierarchical reference structures and hierarchical quality structures have been extensively studied in traditional video coding~\cite{Overview_of_SHVC_AVC,Overview_of_SHVC_HEVC} and fidelity-oriented NVC~\cite{ehvc,dcvc-fm}, where they improve temporal prediction, reduce error propagation, and enhance rate-distortion (RD) performance. This motivates hierarchical design in diffusion-based GVC, so that important references can be better preserved during coding and more reliably reused during generative reconstruction.

To address these limitations, we propose a \textbf{G}enerative \textbf{V}ideo \textbf{C}ompression framework based on \textbf{Hi}erarchical \textbf{R}eferencing, named \textbf{\ours}. The key idea is to organize latent frames according to their reference importance, so that a small set of more reliable frames can better support both latent coding and generative reconstruction. To this end, \ours couples a hierarchical reference structure with a hierarchical quality structure in latent coding, and carries this hierarchical structure over to generative reconstruction through hierarchy-aware conditioning.

In latent coding, \ours organizes latent frames into multiple layers. Lower-layer latent frames are allocated more bits and reused more often as references, whereas higher-layer latent frames are coded more compactly and cannot serve as references for lower-layer ones. Built on this design, we introduce \textbf{H}ierarchical \textbf{T}emporal \textbf{C}ontext \textbf{M}ining (\textbf{HTCM}) to exploit complementary cues from short- and long-term references. The short-term reference preserves local details from the nearest decoded latent frame at the same or lower layers, while the long-term reference summarizes global temporal semantics from previously decoded lowest-layer latent frames. HTCM then aligns and fuses these two forms of context to improve conditional coding under hierarchical referencing.

In generative reconstruction, \ours injects decoded latent frames into a video diffusion transformer (DiT) through a \textbf{H}ierarchical \textbf{A}ttentive \textbf{Adapter} (\textbf{HA-Adapter}) to guide denoising~\cite{vace}. Instead of using full attention across all frames, the HA-Adapter preserves the coding-side hierarchy through hierarchical attention, so that each latent frame attends only to the same- or lower-layer references. This prevents less reliable higher-layer latent frames, coded with fewer bits, from contaminating more important lower-layer references during denoising, thereby reducing artifact propagation in generative reconstruction.

To the best of our knowledge, this work is the first unified framework in diffusion-based GVC that extends hierarchical design from latent coding to generative reconstruction. This design offers two advantages: 1) it allocates more bits to latent frames with more important reference roles during coding, thereby reducing the overall bitrate; and 2) it enables the diffusion model to rely on these higher-quality references to reconstruct the remaining frames more effectively, thus improving perceptual quality under aggressive compression. Experiments demonstrate that \ours achieves state-of-the-art compression performance across multiple metrics. Compared with the latest GVC method~\cite{gnvc-vd}, \ours achieves 50.5\% and 54.0\% BD-rate reductions in terms of LPIPS and DISTS, respectively, while also producing better visual results, as shown in Fig.~\ref{fig:videoSRC24}.

Our contributions are summarized as follows:

\begin{itemize}
  \item We propose \ours, a generative video compression framework based on hierarchical referencing, which is the first to explore hierarchical reference structures for both latent coding and generative reconstruction.
  \item We introduce hierarchical temporal context mining for latent coding, which exploits complementary short- and long-term temporal context from high-quality lower-layer references.
  \item We design a hierarchical attentive adapter to inject decoded latent frames into a video DiT through hierarchical attention, thereby reducing coding artifact propagation during denoising.
  \item \ours achieves state-of-the-art compression performance in diffusion-based GVC, achieving 50.5\% and 54.0\% BD-rate reductions in terms of LPIPS and DISTS, respectively, compared with the latest method~\cite{gnvc-vd}.
\end{itemize}

The rest of this paper is organized as follows. Section~\ref{sec:related work} reviews GVC and hierarchical reference structures. Section~\ref{sec:methodology} presents \ours. Experimental results are given in Section~\ref{sec:experiments}, and Section~\ref{sec:conclusion} concludes the paper.

\begin{figure*}[t]
\centering
\includegraphics[width=\textwidth]{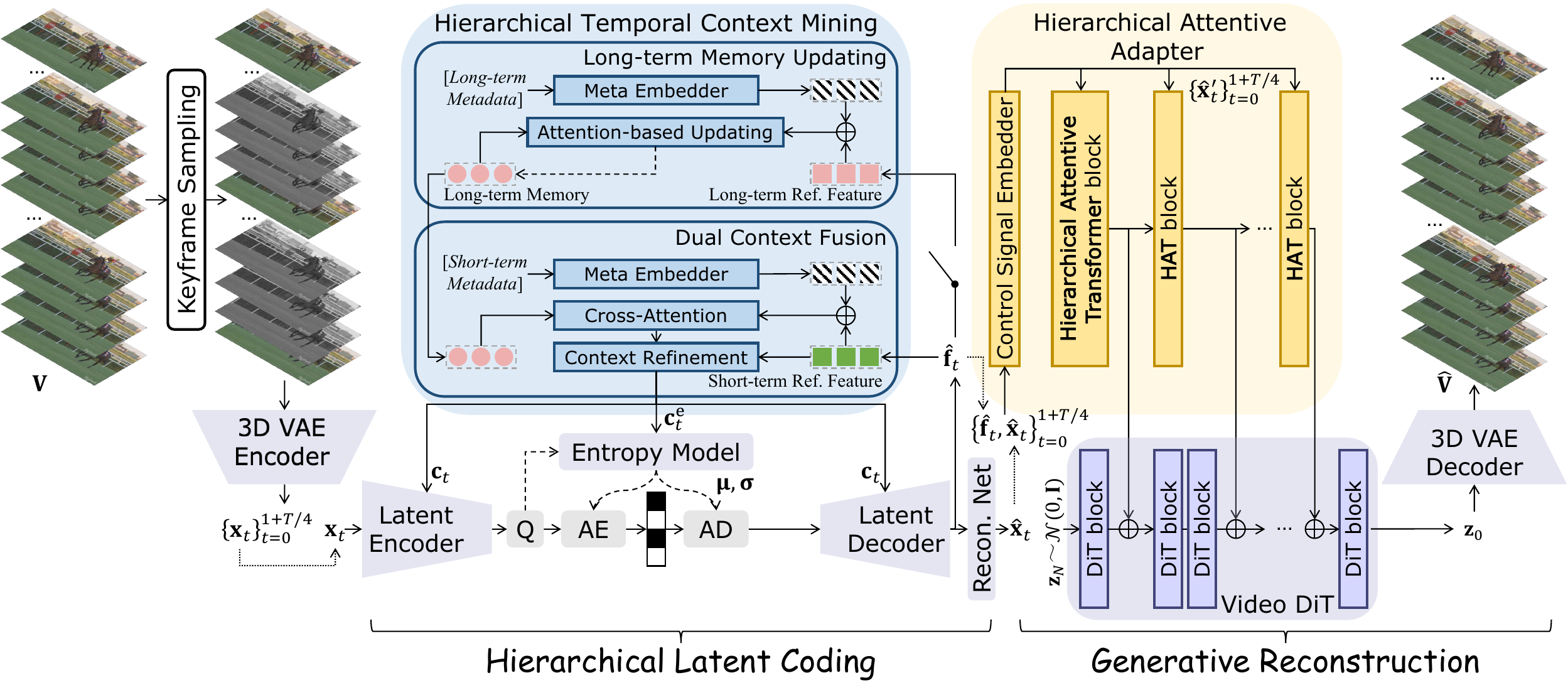}
\vspace{-25pt}
\caption{Overview of the proposed GVCHR framework for generative video compression.}
\vspace{-10pt}
\label{fig:framework}
\end{figure*}

\section{Related Work}\label{sec:related work}
\subsection{Generative Video Compression}
Generative video compression (GVC) exploits generative models to improve reconstruction and restore realistic details, thereby improving perceptual quality beyond the pixel fidelity targeted by traditional video coding and fidelity-oriented NVC. Early studies~\cite{dvc-p,NVCusingGAN,plvc,hvfvc} mainly incorporated generative adversarial networks (GANs) and perceptual losses into existing neural codecs. For example, PLVC~\cite{plvc} adopted a recurrent autoencoder together with a recurrent discriminator under GAN training. However, such methods often suffer from training instability and visually unpleasant artifacts caused by mode collapse.

Subsequent studies explored latent-space generation for video reconstruction. GLC-Video~\cite{glc} performed transform coding in the latent space of a vector-quantized variational autoencoder (VQ-VAE)~\cite{vq-vae}, but its reconstruction quality was limited by the representational capacity of VQ-VAE. More recently, diffusion models have become popular due to their strong ability to synthesize realistic spatial and temporal details. I$^2$VC~\cite{i2vc} utilized frame-level latent diffusion to align reference features and generate target frames by denoising motion-related regions. DiffVC~\cite{diffvc} compressed videos in the latent domain and then used an image diffusion model to enhance decoded latent frames. Other approaches~\cite{m3-cvc,t-gvc,diffusion-aided} improved compression efficiency by transmitting keyframes and limited side information, while using a video diffusion model to generate the remaining frames at the decoder.

Despite their promising perceptual quality, these methods still exhibit limitations. Methods based on image diffusion priors may suffer from temporal flickering, while those relying on sparse reconstruction cues, such as keyframes and lightweight side information, often compromise content fidelity. To alleviate these issues, GNVC-VD~\cite{gnvc-vd} employed a neural video codec to encode latent frames and used a video DiT to jointly enhance decoded latents within a group of pictures (GOP). Nevertheless, existing diffusion-based GVC methods have not explicitly explored hierarchical organization of latent frames with respect to their reference roles and coding quality, nor have they preserved coding-side reference structures in generative reconstruction.

\subsection{Hierarchical Reference Structure}
Hierarchical reference structures organize frames in a GOP into multiple layers, where lower-layer frames are reused more often as references for higher-layer ones. In traditional video coding, such structures are typically combined with hierarchical quality structures, where lower-layer frames are encoded with finer quantization and higher fidelity~\cite{Analysis_of_hierarchical_B_pictures}. Compared with simple sequential reference chains, hierarchical reference structures reduce error propagation, improve RD performance, and support temporal scalability~\cite{Overview_of_SHVC_HEVC,Overview_of_SHVC_AVC}.

Hierarchical designs have been widely studied in NVC. Under low-delay settings, early work explored hierarchical quality structures for frame-level bit allocation in end-to-end RD optimization. DCVC-DC~\cite{dcvc-dc} scaled the distortion term using hierarchy-dependent frame weights, and DCVC-FM~\cite{dcvc-fm} periodically refreshed reference features to suppress error propagation over long prediction chains. Beyond quality variation alone, DVCH~\cite{dvch} aligned hierarchical reference and quality structures under the low-delay B setting, while EHVC~\cite{ehvc} introduced a hierarchical multi-reference scheme with learnable layer-wise quantization scales. Under random-access settings, HLVC~\cite{hlvc} organized a GOP into quality layers and recurrently enhanced higher-layer frames using bidirectional references from lower layers. Later approaches~\cite{b-canf,lhbdc,tlzmc,dcvc-b,biecvc,brhvc} further explored bidirectional references or context to improve temporal prediction.

Yet, hierarchical design has been explored only partially in diffusion-based GVC. DiffVC~\cite{diffvc} adopted a simple frame-by-frame reference chain together with a hierarchical quality structure inherited from~\cite{dcvc-dc}, while GNVC-VD~\cite{gnvc-vd} also relied on frame-by-frame referencing and allocated bits uniformly across latent frames. Existing methods do not jointly organize reference structures and quality allocation in latent coding, nor do they carry coding-side hierarchical structure over to generative reconstruction. As a result, a unified hierarchical design across coding and generation remains unexplored in diffusion-based GVC.

\section{Methodology}\label{sec:methodology}

\begin{figure*}[t]
  \centering

  \begin{minipage}[c]{0.325\textwidth}
    \centering
    \subfloat[Previous frame-by-frame referencing]{
      \includegraphics[width=\linewidth]{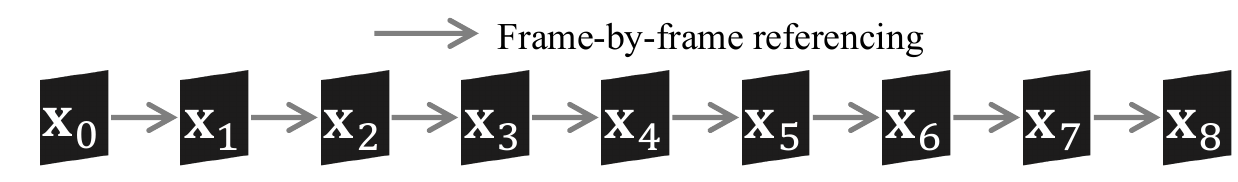}
      \label{subfig:frame_by_frame_reference_in_coding}
    }

    \vspace{-5pt}

    \subfloat[Our hierarchical referencing with two layers.]{
      \includegraphics[width=\linewidth]{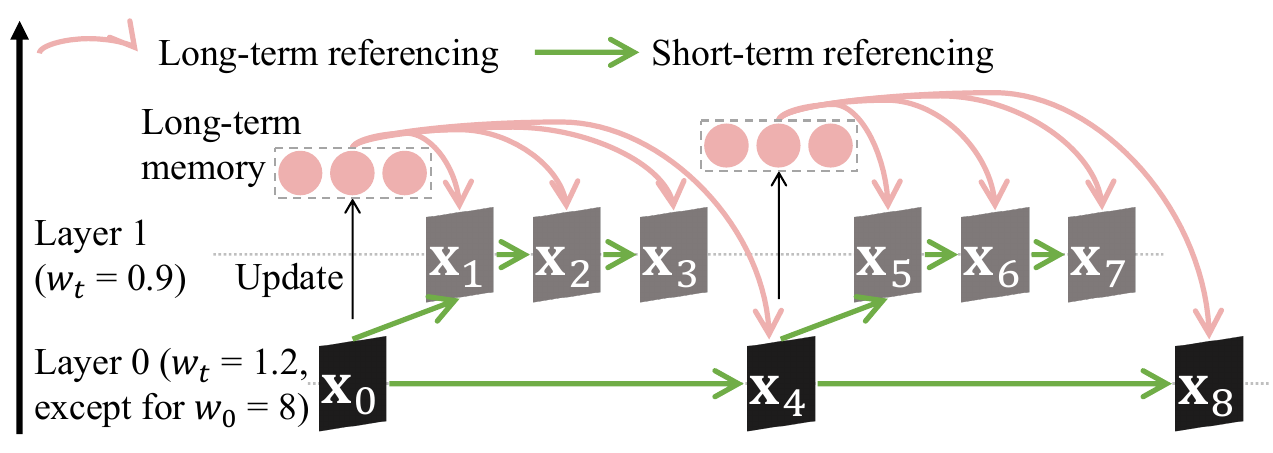}
      \label{subfig:hierarchies_2_coding}
    }
  \end{minipage}
  \hfill
  \begin{minipage}[c]{0.325\textwidth}
    \centering
    \subfloat[Our hierarchical referencing with three layers.]{
      \includegraphics[width=\linewidth]{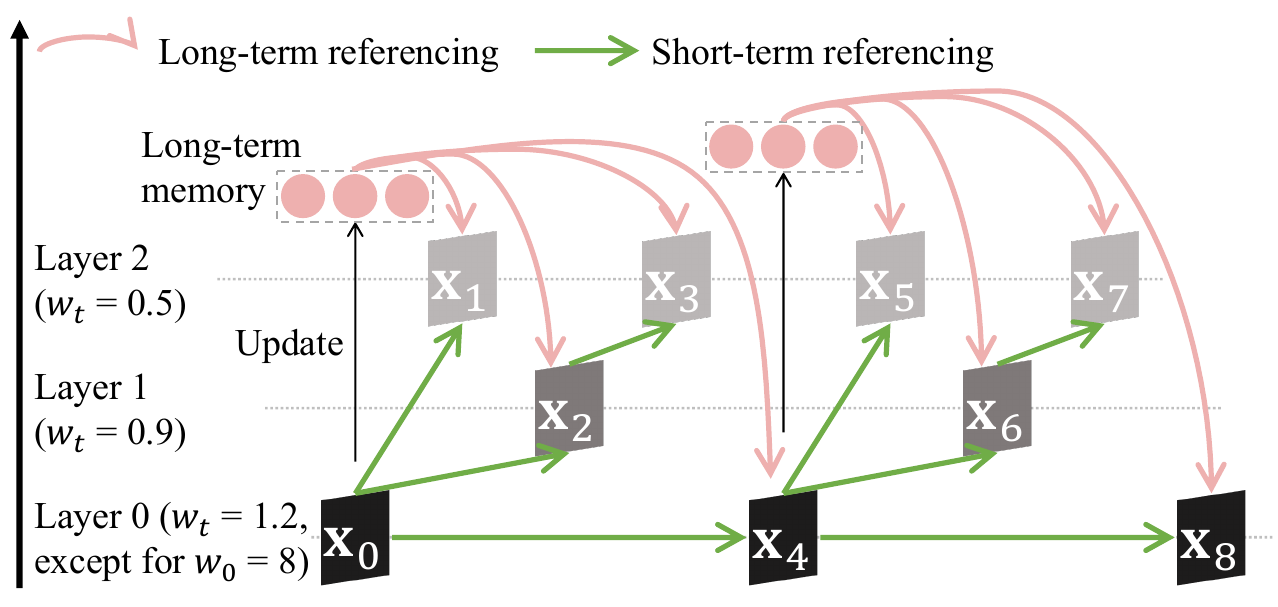}
      \label{subfig:hierarchies_3_coding}
    }
  \end{minipage}
  \hfill
  \begin{minipage}[c]{0.325\textwidth}
    \centering
    \subfloat[Our hierarchical referencing with four layers.]{
      \includegraphics[width=\linewidth]{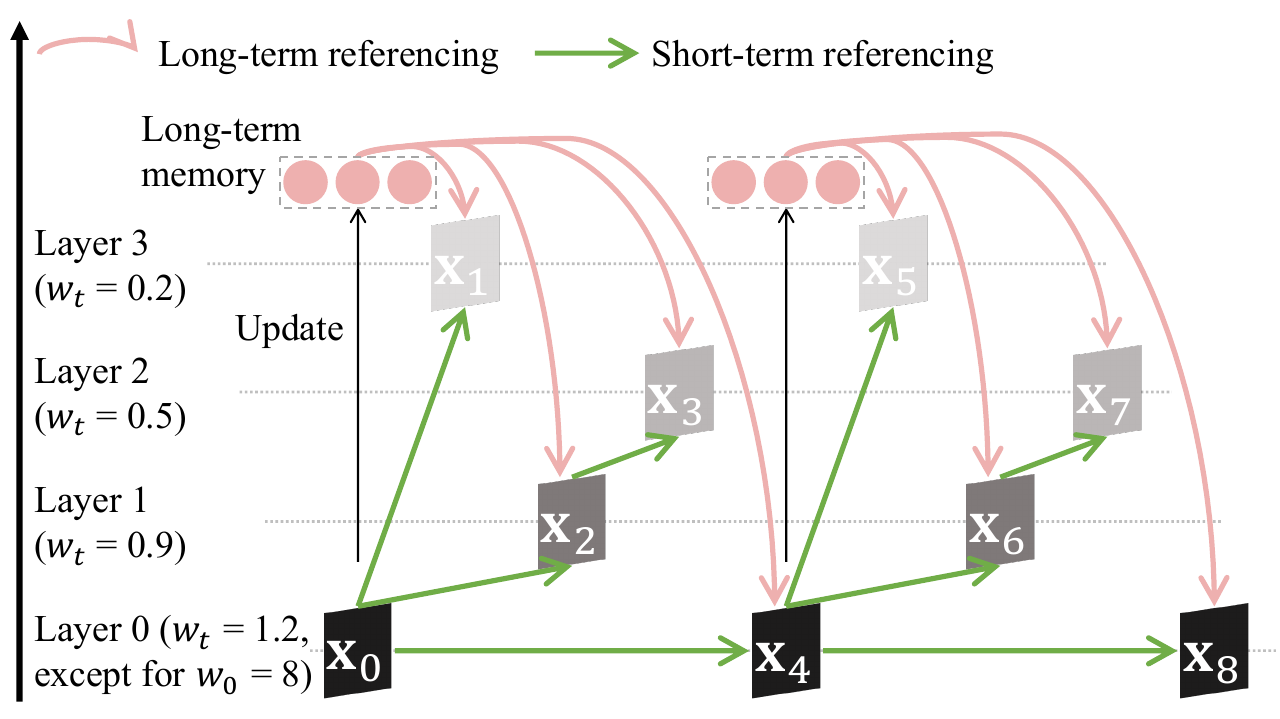}
      \label{subfig:hierarchies_4_coding}
    }
  \end{minipage}

  \caption{Different settings of hierarchical reference structures. Wherein, $w_t$ determines the frame-level coding quality and is employed in the RD loss.}
  \label{fig:hierarchies_in_coding}
  \vspace{-10pt}
\end{figure*}

\subsection{Overview}
In this work, we aim to improve perceptual video reconstruction in diffusion-based GVC through a unified hierarchical design across coding and generation. To this end, \ours organizes latent frames with coupled hierarchical reference and quality structures in latent coding, and preserves the resulting hierarchy during generative reconstruction. As illustrated in Fig.~\ref{fig:framework}, \ours consists of hierarchical latent coding and diffusion-based generative reconstruction. The remainder of this section first introduces the overall framework, followed by Hierarchical Temporal Context Mining for latent coding in Section~\ref{subsec:htcm}, the Hierarchical Attentive Adapter for generative reconstruction in Section~\ref{subsec:ha-adapter}, and the multi-stage training scheme in Section~\ref{subsec:multi-stage training strategies}.

\subsubsection{Hierarchical Latent Coding}

Source video $\mathbf{V}\in \mathbb{R}^{(1+T)\times H\times W \times3}$ is first separated into keyframes and grayscale intermediate frames. The discarded color information of intermediate frames is later recovered during generation, which introduces little degradation in reconstruction quality while reducing the bitrate required for transmission. The resulting visual signals are then encoded to latent frames $\{\mathbf{x}_t \}_{t=0}^{1+T/4}$ by a 3D VAE encoder~\cite{vace}. 

To remove redundancy within $\{\mathbf{x}_t \}_{t=0}^{1+T/4}$, each latent frame $\mathbf{x}_t$ is compressed by a latent codec built upon the conditional coding framework in~\cite{dcvc-rt}. The codec is enhanced with the proposed Hierarchical Temporal Context Mining (HTCM), which exploits a hierarchical reference structure to extract temporal context from previously decoded features: 
\begin{align}
    \mathbf{y}_t &= g_\mathrm{a}(\mathbf{x}_t \mid \mathbf{c}_t),\\
    \hat{\mathbf{y}}_t &= \lfloor \mathbf{y}_t \rceil, \\
    \hat{\mathbf{x}}_t, \hat{\mathbf{f}}_t &= g_\mathrm{s}(\hat{\mathbf{y}}_t \mid \mathbf{c}_t), \\
    \boldsymbol{\upmu}, \boldsymbol{\upsigma} &= g_\mathrm{ep}(\mathbf{y}_t \mid \mathbf{c}_t^\mathrm{e}), \\
    \mathbf{c}_t, \mathbf{c}_t^\mathrm{e} &= \operatorname{HTCM}(\hat{\mathbf{f}}_{<t}),
\end{align}
where $g_\mathrm{a}$ and $g_\mathrm{s}$ denote the latent encoder and decoder, respectively, and $\lfloor \cdot \rceil$ denotes rounding quantization. The decoded latent frame $\hat{\mathbf{x}}_t$ and its feature $\hat{\mathbf{f}}_t$ are reconstructed from $\hat{\mathbf{y}}_t$. The entropy model $g_\mathrm{ep}$ estimates the parameters $\boldsymbol{\upmu}$ and $\boldsymbol{\upsigma}$ of a Gaussian probability model for entropy coding. HTCM takes the previously decoded features $\hat{\mathbf{f}}_{<t}$ as input and produces two temporal context representations, where $\mathbf{c}_t$ is used for conditional latent encoding and decoding, and $\mathbf{c}_t^\mathrm{e}$ for entropy parameter estimation. By organizing latent frames hierarchically, HTCM allocates more bits to lower-layer frames reused more frequently as references, and mines temporal context from their decoded features for more effective conditional coding. Details of HTCM are presented in Section~\ref{subsec:htcm}.

\subsubsection{Diffusion-based Generative Reconstruction}
For generative reconstruction, the decoded latent frames $\{\hat{\mathbf{x}}_t\}_{t=0}^{1+T/4}$ and their features $\{\hat{\mathbf{f}}_t\}_{t=0}^{1+T/4}$ are fed into a controllable video generation model composed of a video DiT and the proposed Hierarchical Attentive Adapter (HA-Adapter). We adopt a flow-matching formulation that learns a continuous velocity field $\boldsymbol{\epsilon}_\tau$ transporting Gaussian noise $\mathbf{z}_N\sim \mathcal{N}(0, \mathbf{I})$ toward the data manifold $\mathbf{z}_0$~\cite{flow_matching}. To guide denoising, HA-Adapter fuses $\{\hat{\mathbf{x}}_t\}_{t=0}^{1+T/4}$ and $\{\hat{\mathbf{f}}_t\}_{t=0}^{1+T/4}$ into frame-wise representations $\{\hat{\mathbf{x}}'_t\}_{t=0}^{1+T/4}$ as control signals, which are injected into the video DiT through hierarchical attentive transformer blocks. Here, hierarchical attention restricts each frame to attend only to the same- or lower-layer references, preventing less reliable higher-layer frames from propagating artifacts during denoising. Details of HA-Adapter are given in Section~\ref{subsec:ha-adapter}.

The denoising process is formulated as:
\begin{align}
    \mathbf{z}_{\tau-1}&=\mathbf{z}_{\tau}+(\sigma_{\tau-1}-\sigma_{\tau})\cdot \hat{\epsilon}(\mathbf{z}_{\tau},\tau, \mathbf{u}),\\
    \mathbf{u}&=\operatorname{HA-Adapter}(\{\hat{\mathbf{x}}_t \}_{t=0}^{1+T/4}, \{\hat{\mathbf{f}}_t \}_{t=0}^{1+T/4}),
\end{align}
where $\mathbf{z}_{\tau}\in \mathbb{R}^{(1+T/4)\times(H/8)\times (W/8)\times16}$ denotes the latent variable along the denoising trajectory, and $16$ is the channel dimension of the 3D VAE bottleneck. The index $\tau=0,1,...,N-1$ enumerates the flow steps and $\sigma_{\tau}$ is the noise scale specified by a predefined schedule~\cite{wan}. The video DiT predicts the velocity field $\hat{\epsilon}(\mathbf{z}_{\tau},\tau, \mathbf{u})$, and HA-Adapter projects the decoded latents and their features into the conditioning signal $\mathbf{u}$ aligned with the intermediate feature space of the transformer. After $N$ denoising steps, the generated latent $\mathbf{z}_0$ is decoded by a 3D VAE decoder into the final pixel-space reconstruction $\hat{\mathbf{V}}$.

\subsection{Hierarchical Temporal Context Mining}\label{subsec:htcm}

\begin{figure*}[t]
  \centering
  \subfloat[Previous all-to-all referencing in attention.]{
    \includegraphics[width=0.275\textwidth]{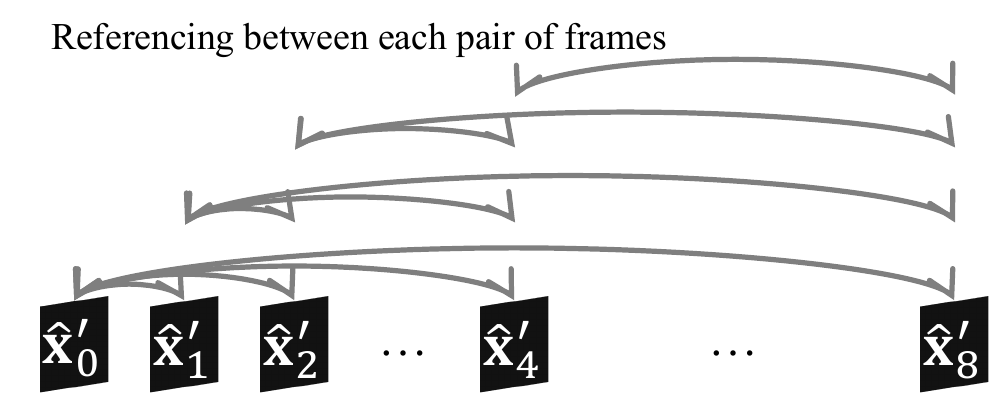}
    \label{subfig:all_to_all_referencing_in_generation}
  }
  \hfill
  \subfloat[Full-attention mask]{
    \includegraphics[width=0.155\textwidth]{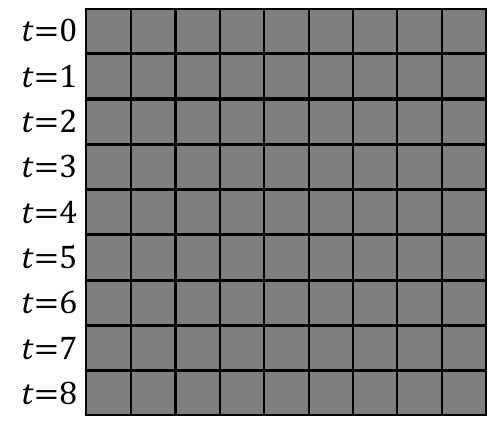}
    \label{subfig:full-attention_mask}
  }
  \hfill
  \subfloat[Our hierarchical referencing in attention.]{
    \includegraphics[width=0.325\textwidth]{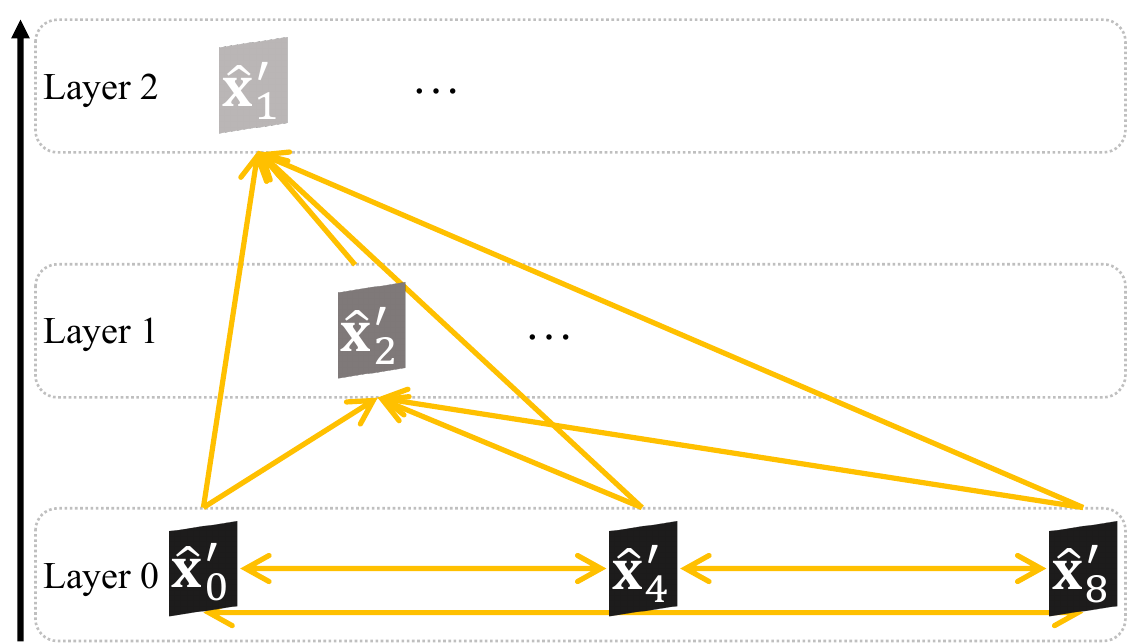}
    \label{subfig:hierarchical_referencing_in_generation}
  }
  \hfill
  \subfloat[Block-attention mask.]{
    \includegraphics[width=0.19\textwidth]{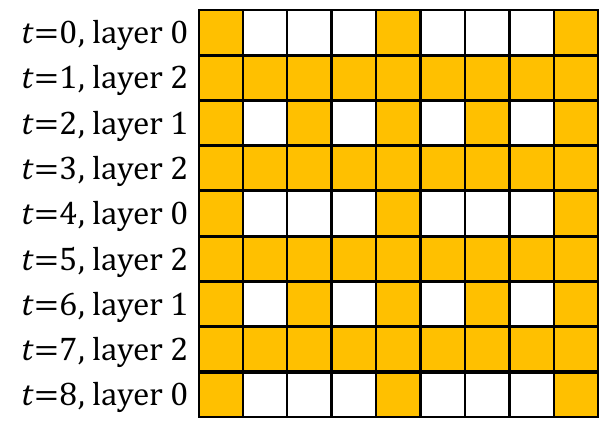}
    \label{subfig:block-wise_attention_mask}
  }
  \caption{Comparison between the previous all-to-all reference structure~\cite{gnvc-vd} and our hierarchical reference structure performed by attention in the adapter. For clarity, $\hat{\mathbf{x}}'_3$, $\hat{\mathbf{x}}'_5$, $\hat{\mathbf{x}}'_6$, and $\hat{\mathbf{x}}'_7$ are omitted in Fig.~\ref{subfig:all_to_all_referencing_in_generation} and Fig.~\ref{subfig:hierarchical_referencing_in_generation}. Each \textcolor[rgb]{1,0.75,0}{yellow} block in Fig.~\ref{subfig:block-wise_attention_mask} corresponds to $\mathbf{M}_{i,j}=\mathbf{0}$ in Equation~\ref{eq: referencing determination in attention mask}, indicating that $\hat{\mathbf{x}}'_j$, selected from $\{\hat{\mathbf{x}}'_t\}_{t=0}^{1+T/4}$, serves as a reference for $\hat{\mathbf{x}}'_i$. 
  }
  \label{fig:reference structure in generation}
  \vspace{-10pt}
\end{figure*}

Informative temporal context is crucial for reducing temporal redundancy in latent coding~\cite{dcvc-tcm}. To this end, we propose HTCM, which exploits the hierarchical reference structure over $\{\mathbf{x}_t\}_{t=0}^{1+T/4}$ to provide more informative temporal conditions for conditional coding. We first describe the hierarchical reference structure used in coding and then present short- and long-term context mining from selected references.

\subsubsection{Hierarchical Reference Structure for Coding}

Under the hierarchical reference structure, $\{\mathbf{x}_t\}_{t=0}^{1+T/4}$ are organized into multiple layers such that higher-layer latent frames cannot serve as references for lower-layer ones. The associated hierarchical quality structure allocates more bits to lower-layer frames, which are reused more often as references. Although higher-layer latent frames are coded more compactly and contain more distortion, their limited use as references mitigates error accumulation and propagation. Such bit allocation is also compatible with keyframe sampling, where lowest-layer latent frames retain keyframe-related information and receive more bits to preserve color fidelity, while higher-layer frames retain only grayscale information.

We investigate two-, three-, and four-layer settings under a GOP size of 32, corresponding to a latent GOP size of 8. The latent mini-GOP refers to a periodically repeated reference and quality pattern of size 4, as shown in Fig.~\ref{fig:hierarchies_in_coding}. The coding quality of $\mathbf{x}_t$ is controlled by $w_t$, a hierarchical temporal weight in the RD training loss, where latent frames at the same layer share the same weight. We denote $L_t$ as the layer ID of $\mathbf{x}_t$. As shown in Section~\ref{subsubsec: ablation on hierarchical reference structure}, the three-layer setting gives the best compression performance. Specifically, for the 4-frame latent mini-GOP starting from the first inter-frame $\mathbf{x}_1$, the weights for $\mathbf{x}_1$ to $\mathbf{x}_4$ are $0.5$, $0.9$, $0.5$, and $1.2$, with corresponding layer IDs $\{L_t\}_{t=1}^{4} = \{2,1,2,0\}$.

\subsubsection{Short- and Long-term Context Mining}
Based on the hierarchical reference structure, HTCM extracts complementary temporal context from short- and long-term references, capturing local details and global semantics, respectively. For each latent frame $\mathbf{x}_t$, the short-term reference is the feature of the nearest previously decoded latent frame in the same or lower layers, denoted by $\hat{\mathbf{f}}_{s}$, where $s = \max \{ i \mid i < t, L_i \le L_t \}$. As a carrier of local details, $\hat{\mathbf{f}}_s$ is used as the short-term context:
\begin{equation}
   \mathbf{c}_t^{\mathrm{S}} \coloneqq \hat{\mathbf{f}}_s.
\end{equation}
In contrast, the long-term references consist of all previously decoded lowest-layer frames, represented by their corresponding features $\{\hat{\mathbf{f}}_l \mid l < t, L_l = 0\}$. After each feature $\hat{\mathbf{f}}_l$ is decoded, we use gated slot attention (GSA)~\cite{gsa} to recurrently update a fixed-size memory and obtain the corresponding retrieved long-term representation $\mathbf{m}_l$:
\begin{align}
    \mathbf{m}_{l} &= \operatorname{GSA}(\mathbf{q}_{l},\mathbf{k}_{l},\mathbf{v}_{l},\boldsymbol{\upalpha}_{l}),\\
    \mathbf{q}_{l},\mathbf{k}_{l},\mathbf{v}_{l}&= \mathbf{W}_\mathrm{q}^\mathrm{G}(\hat{\mathbf{f}}_{l}+\mathbf{e}_l), \mathbf{W}_\mathrm{k}^\mathrm{G}(\hat{\mathbf{f}}_{l}+\mathbf{e}_l), \mathbf{W}_\mathrm{v}^\mathrm{G}(\hat{\mathbf{f}}_{l}+\mathbf{e}_l),\\
    \boldsymbol{\upalpha}_{l}&= \mathbf{W}_\upalpha^\mathrm{G}(\hat{\mathbf{f}}_{l}+\mathbf{e}_l),\\
    \mathbf{e}_l &= \operatorname{Linear}(\operatorname{SiLU}(\operatorname{Linear}([L_l,l,\mathrm{qp}_l]))),
\end{align}
where $\mathbf{W}_\mathrm{q}^\mathrm{G}$, $\mathbf{W}_\mathrm{k}^\mathrm{G}$, $\mathbf{W}_\mathrm{v}^\mathrm{G}$, and $\mathbf{W}_\upalpha^\mathrm{G}$ are learnable projection matrices. $\mathbf{q}_l$, $\mathbf{k}_l$, and $\mathbf{v}_l$ denote the query, key, and value, respectively, and $\boldsymbol{\upalpha}_l$ is a data-dependent gating vector. The meta embedding $\mathbf{e}_l$ encodes the layer ID $L_l$, latent frame index $l$, and quantization parameter $\mathrm{qp}_l$. The GSA update and long-term representation retrieval are formulated as: 
\begin{align}
    \mathbf{m}_l &= (\mathbf{h}^{\mathrm{v}}_\mathrm{cur})^\top \operatorname{softmax}\left((\mathbf{h}^{\mathrm{k}}_\mathrm{cur})^\top \mathbf{q}_l\right),\\
    \mathbf{h}^{\mathrm{k}}_\mathrm{cur} &= \operatorname{Diag}\left(\boldsymbol{\upalpha}_l\right)\cdot \mathbf{h}^{\mathrm{k}}_\mathrm{pre}+\left(1-\boldsymbol{\upalpha}_l\right)\otimes \mathbf{k}_l, \\
    \mathbf{h}^{\mathrm{v}}_\mathrm{cur} &= \operatorname{Diag}\left(\boldsymbol{\upalpha}_l\right)\cdot \mathbf{h}^{\mathrm{v}}_\mathrm{pre}+\left(1-\boldsymbol{\upalpha}_l\right)\otimes \mathbf{v}_l,
\end{align}
where $\mathbf{h}^{\mathrm{k}}_\mathrm{cur}$ and $\mathbf{h}^{\mathrm{v}}_\mathrm{cur}$ denote the memory states at the current step, updated by the long-term reference $\hat{\mathbf{f}}_l$. $\mathbf{h}^{\mathrm{k}}_\mathrm{pre}$ and $\mathbf{h}^{\mathrm{v}}_\mathrm{pre}$ denote the states from the previous update step. $\otimes$ denotes the outer product.

For the current frame $\mathbf{x}_t$, we use the latest available long-term representation to connect global semantics with local details. Specifically, $\mathbf{m}_l$ is aligned with the short-term context $\mathbf{c}_t^{\mathrm{S}}$ through cross-attention~\cite{attention} to produce the long-term context $\mathbf{c}_t^{\mathrm{L}}$:
\begin{align}
    \mathbf{c}_t^{\mathrm{L}} &= \operatorname{Attention}(\mathbf{W}_\mathrm{q}^\mathrm{C}(\mathbf{c}_{t}^\mathrm{S}+\mathbf{e}_s), \mathbf{W}_\mathrm{k}^\mathrm{C}\mathbf{m}_{l}, \mathbf{W}_\mathrm{v}^\mathrm{C}\mathbf{m}_{l}),\\
    \mathbf{e}_s &= \operatorname{Linear}(\operatorname{SiLU}(\operatorname{Linear}([L_s,s,\mathrm{qp}_s]))),
\end{align}
where $\mathbf{W}_\mathrm{q}^\mathrm{C}$, $\mathbf{W}_\mathrm{k}^\mathrm{C}$, and $\mathbf{W}_\mathrm{v}^\mathrm{C}$ are learnable projection matrices, and $\mathbf{e}_s$ is the meta embedding of $\hat{\mathbf{f}}_s$. Finally, $\mathbf{c}_t^{\mathrm{S}}$ and $\mathbf{c}_t^{\mathrm{L}}$ are fused by a context refinement network, which follows the same architecture as the context extractor in~\cite{dcvc-rt} except that the input channel dimension is doubled, to produce $\mathbf{c}_t$ and $\mathbf{c}_t^\mathrm{e}$ for conditional coding.

\subsection{Hierarchical Attentive Adapter}\label{subsec:ha-adapter}

Once $\{\hat{\mathbf{x}}_t\}_{t=0}^{1+T/4}$ and $\{\hat{\mathbf{f}}_t\}_{t=0}^{1+T/4}$ are decoded, the proposed HA-Adapter injects them into a video DiT as control signals for generative reconstruction. HA-Adapter is based on the adapter in VACE~\cite{vace} with the original attention replaced by hierarchical attention. In this section, we first describe the hierarchical reference structure for generation, and then detail its implementation in hierarchical attentive transformer (HAT) blocks.

\subsubsection{Hierarchical Reference Structure for Generation}

As illustrated in Fig.~\ref{fig:framework}, we first employ a control signal embedder, where $\{\hat{\mathbf{x}}_t\}_{t=0}^{1+T/4}$ and $\{\hat{\mathbf{f}}_t\}_{t=0}^{1+T/4}$ are concatenated and patchified into frame-wise tokens $\{\hat{\mathbf{x}}'_t\}_{t=0}^{1+T/4}$. Patchification is implemented using a convolutional layer with spatial downsampling factor of 2, such that 
$\hat{\mathbf{x}}'_t \in \mathbb{R}^{(H/16 \cdot W/16)\times d}$, where $d$ is the token dimension. To project $\{\hat{\mathbf{x}}'_t\}_{t=0}^{1+T/4}$ into the feature space of the video DiT, GNVC-VD~\cite{gnvc-vd} uses full attention within transformer blocks to learn spatiotemporal correlations among all frame-wise tokens. In contrast, we use hierarchical attention to construct hierarchical reference structure for generation, as shown in Fig.~\ref{fig:reference structure in generation}. Specifically, each frame's tokens attend only to other frames' tokens from the same or lower layers according to the coding hierarchy. Since quality variation across $\{\hat{\mathbf{x}}'_t\}_{t=0}^{1+T/4}$ is inherited from the hierarchical quality structure in coding, this design prevents the propagation of coding artifacts from low-quality frames to high-quality ones. At the same time, each frame can draw cleaner guidance from higher-quality references, thereby reducing artifact propagation. Therefore, the hierarchical structures in coding and generation are intrinsically coupled.
 
\begin{table*}[t]
\caption{Detailed configurations of the multi-stage training strategies.}
\centering
\vspace{-5pt}
\renewcommand{\arraystretch}{1.25}
\renewcommand{\tabcolsep}{1.5pt}
\scalebox{0.925}{
\begin{tabular}{c|c|c|c|c|c|c|c|c}
\toprule
Stage & Iterations & Batch Size & Patch Size & Frames & Learning
Rate & Trainable Models & $\lambda$ & Distortion Term $D$ in the RD loss $L_\mathrm{RD}$ \\ 
\midrule
1 & 30K & 8 & 256 $\times$ 256 & 5 & $10^{-4}$ & Latent Codec & 3072 & $L_\mathrm{Codec}$ \\ 
2 & 20K & 8 & 256 $\times$ 256 & 17 & $10^{-4}$ & Latent Codec & 3072 & $L_\mathrm{Codec}$ \\
3 & 20K & 8 & 256 $\times$ 256 & 33 & $10^{-4}$ & Latent Codec & 3072 & $L_\mathrm{Codec}$ \\
4 & 30K & 8 & 256 $\times$ 256 & 33 & $10^{-4}$ & Latent Codec, HA-Adapter & 3072 &  $L_\mathrm{Codec}+L_\mathrm{CFM}$ \\ 
5 & 40K & 8 & 256 $\times$ 256 & 33 & $5\times 10^{-5}$ & Latent Codec, HA-Adapter & 160, 289, 635, 1397, 3072 & $0.1L_\mathrm{Codec}+0.1L_\mathrm{CFM}+L_\mathrm{Rec}+0.1L_\mathrm{LPIPS}$ \\ 
6 & 60K & 4 & 448 $\times$ 448 & 33 & $2\times 10^{-5}$ & Latent Codec, HA-Adapter & 160, 289, 635, 1397, 3072 & $0.025L_\mathrm{Codec}+L_\mathrm{Rec}+0.1L_\mathrm{LPIPS}$ \\ 
\bottomrule
\end{tabular}
}
\label{tab:training strategy}
\vspace{-10pt}
\end{table*}

\subsubsection{Hierarchical Attentive Transformer}
To enable hierarchical referencing within the adapter, we feed $\{\hat{\mathbf{x}}'_t\}_{t=0}^{1+T/4}$ into HAT blocks instead of the original transformer blocks. The key distinction lies in the incorporation of a block-wise attention mask $\mathbf{M}\in \mathbb{R}^{(H/16 \cdot W/16  \cdot(1+T/4)) \times (H/16\cdot W/16 \cdot(1+T/4))}$:
\begin{align}
\mathbf{o}&=\operatorname{softmax}\left(\frac{\mathbf{q}\mathbf{k}^\top}{\sqrt{d}} + \mathbf{M}\right)\mathbf{v},\\
    \mathbf{M} &=
    \begin{bmatrix}
    \mathbf{M}_{0,0} & \mathbf{M}_{0,1} & \cdots & \mathbf{M}_{0,1+T/4} \\
    \mathbf{M}_{1,0} & \mathbf{M}_{1,1} & \cdots & \mathbf{M}_{1,1+T/4} \\
    \vdots & \vdots & \ddots & \vdots \\
    \mathbf{M}_{1+T/4,0} & \mathbf{M}_{1+T/4,1} & \cdots & \mathbf{M}_{1+T/4,1+T/4}
    \end{bmatrix}, \\
    \mathbf{M}_{i,j} &= \begin{cases}
                         \boldsymbol{-\infty} & \text{if } L_i < L_j \\
                         \mathbf{0} & \text{if } L_i \ge L_j
                        \end{cases},\ 0\le i,j \le 1+T/4,\label{eq: referencing determination in attention mask}
\end{align}
where $\mathbf{q}, \mathbf{k}, \mathbf{v} \in \mathbb{R}^{(H/16 \cdot W/16 \cdot(1+T/4))\times d}$ denote the projected embeddings of $\{\hat{\mathbf{x}}'_t\}_{t=0}^{1+T/4}$ for hierarchical attention in each HAT block. Each block $\mathbf{M}_{i,j} \in \mathbb{R}^{(H/16 \cdot W/16) \times (H/16\cdot W/16)}$ controls the attention weights between $\hat{\mathbf{x}}'_i$ and $\hat{\mathbf{x}}'_j$, whose layer IDs are $L_i$ and $L_j$, respectively. When $\mathbf{M}_{i,j}=\mathbf{0}$, $\hat{\mathbf{x}}'_i$ is allowed to attend to $\hat{\mathbf{x}}'_j$ under the condition $L_i \ge L_j$. Otherwise, $\mathbf{M}_{i,j}=\boldsymbol{-\infty}$ mask out the corresponding attention weights. We retain $\mathbf{M}_{i,j}=\mathbf{0}$ when $i=j$, allowing the model to balance the referenced information and self-information. Fig.~\ref{subfig:block-wise_attention_mask} shows the mask under the three-layer setting. 

Additionally, owing to the block-wise mask, the hierarchical attention in HAT can be accelerated via block attention~\cite{big_bird}, which further reduces the runtime of generative reconstruction.

\subsection{Multi-stage Training Strategies}\label{subsec:multi-stage training strategies}

\begin{figure*}[t]
  \centering
  \subfloat{
    \includegraphics[width=0.245\textwidth]{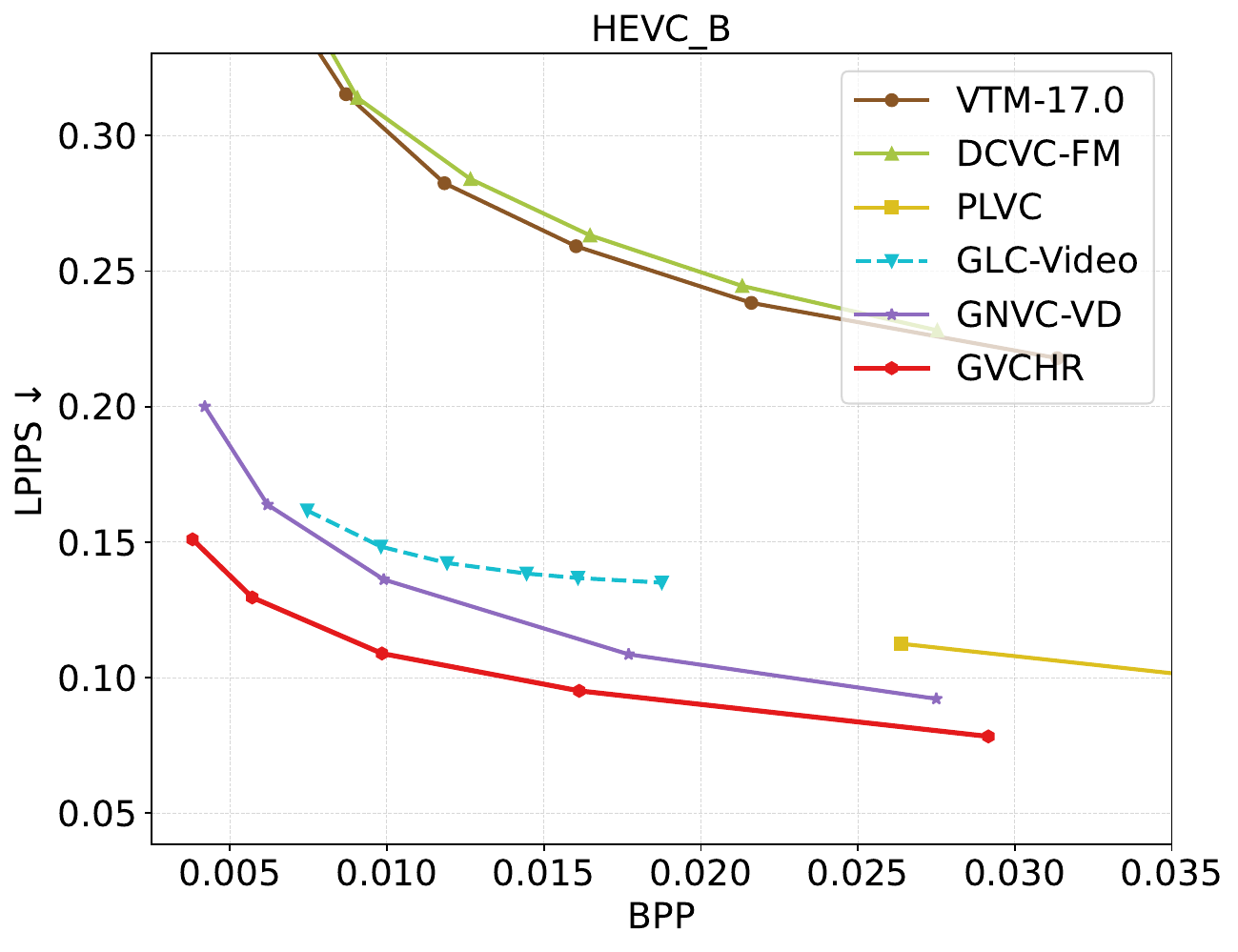}
    \label{subfig:lpips_HEVC_B}
  }
  \hspace{-10pt}
  \subfloat{
    \includegraphics[width=0.245\textwidth]{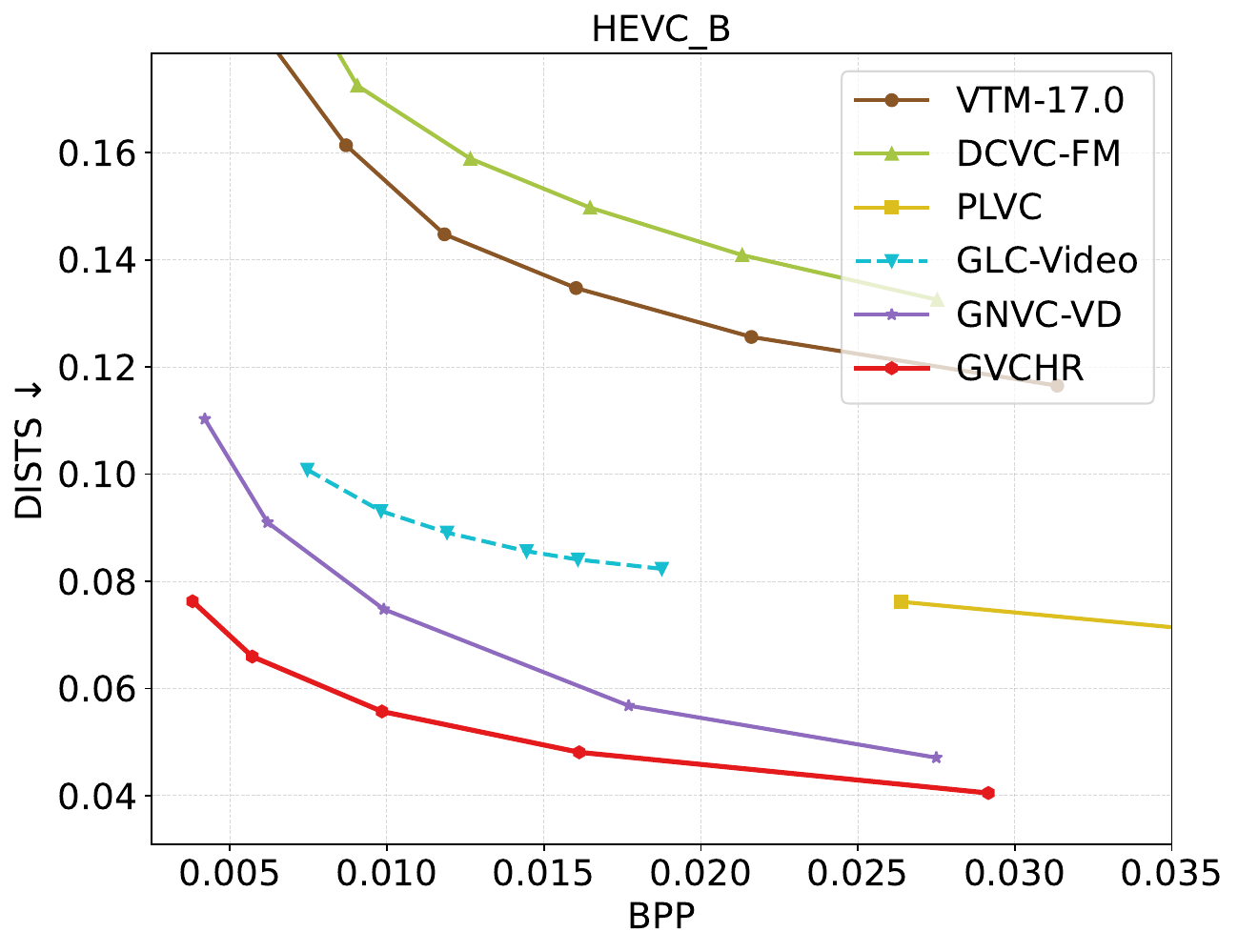}
    \label{subfig:dists_HEVC_B}
  }
  \hspace{-10pt}
  \subfloat{
    \includegraphics[width=0.245\textwidth]{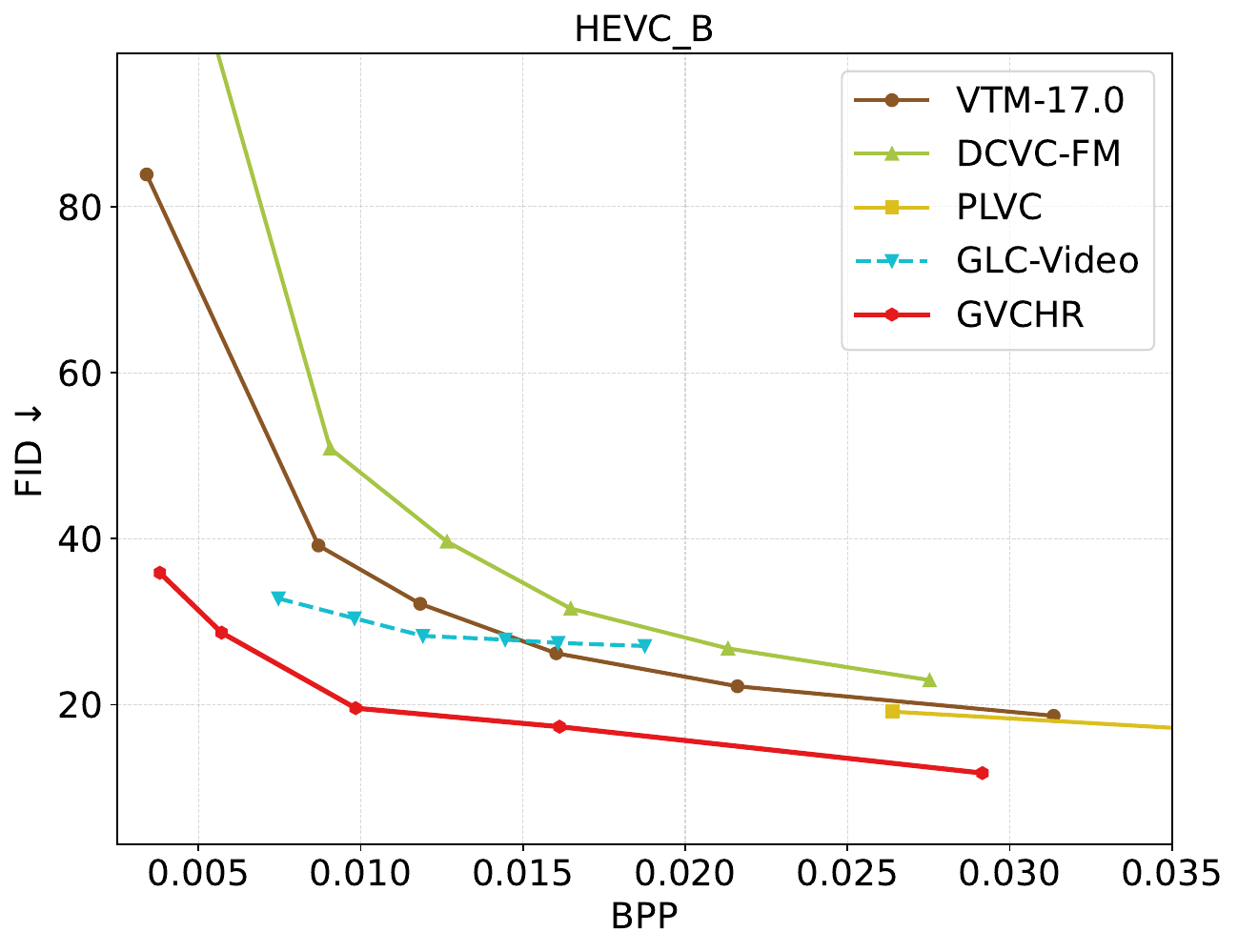}
    \label{subfig:fid_HEVC_B}
  }
  \hspace{-10pt}
  \subfloat{
    \includegraphics[width=0.245\textwidth]{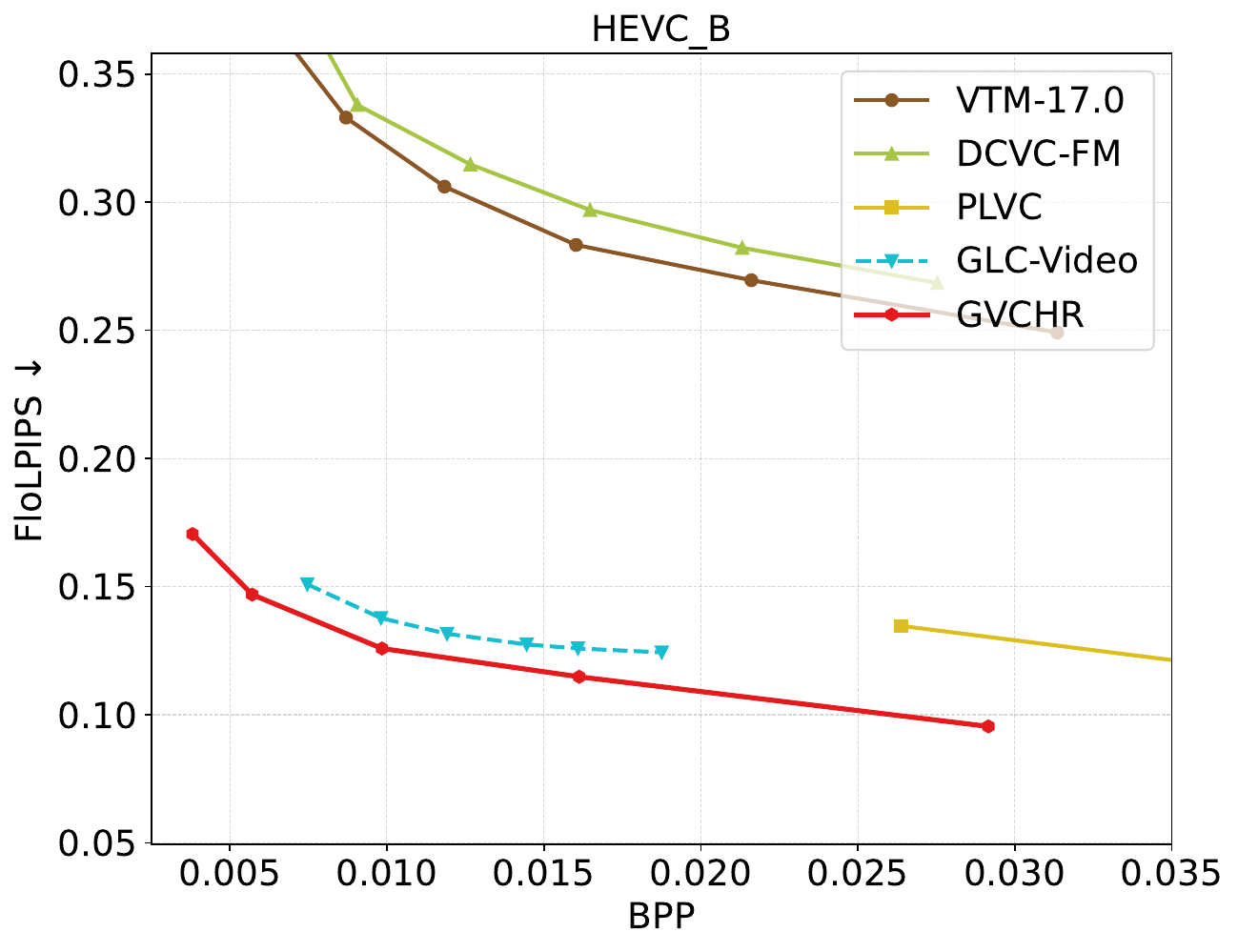}
    \label{subfig:flolpips_HEVC_B}
  }\\[-10pt]
  
  \subfloat{
    \includegraphics[width=0.245\textwidth]{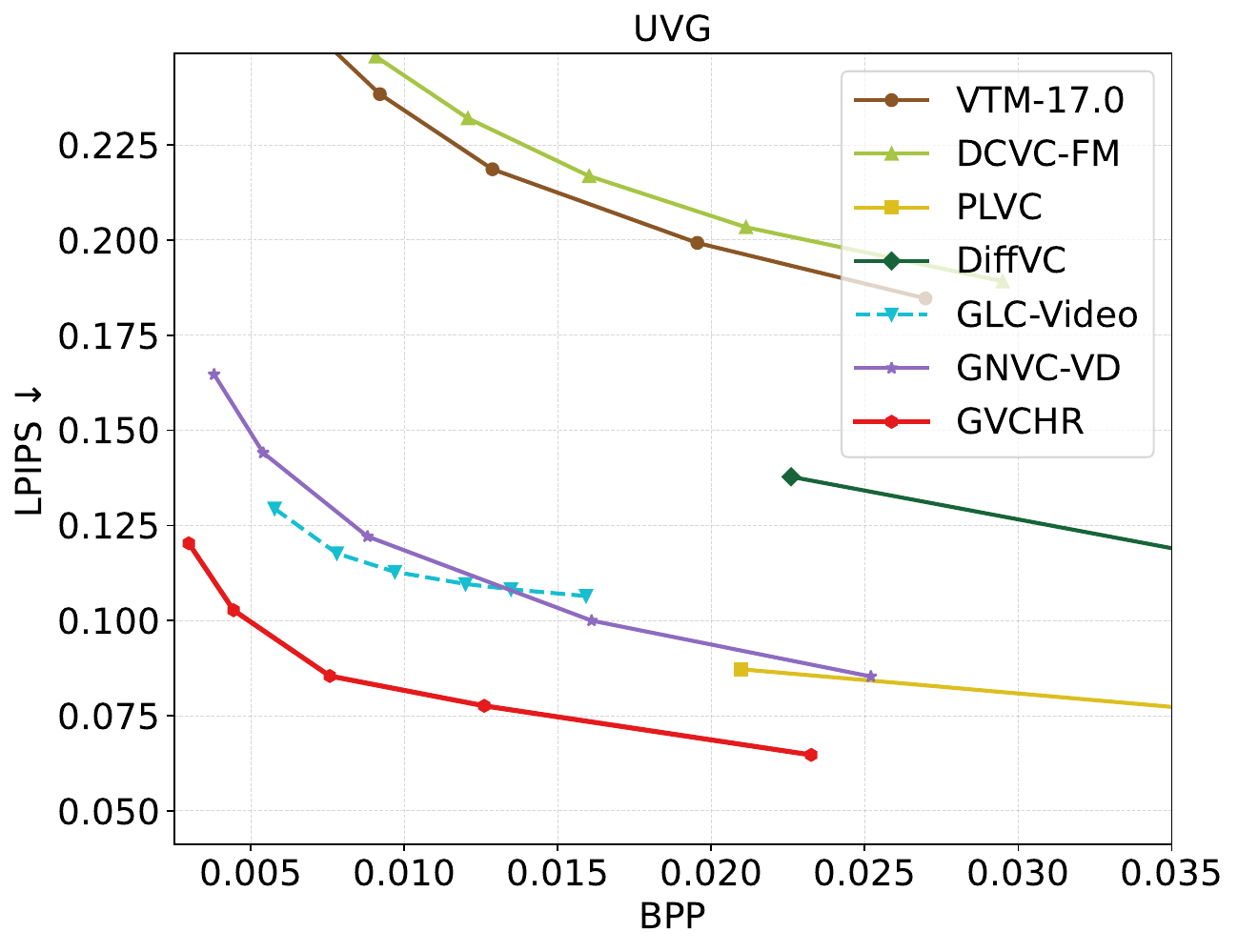}
    \label{subfig:lpips_UVG}
  }
  \hspace{-10pt}
  \subfloat{
    \includegraphics[width=0.245\textwidth]{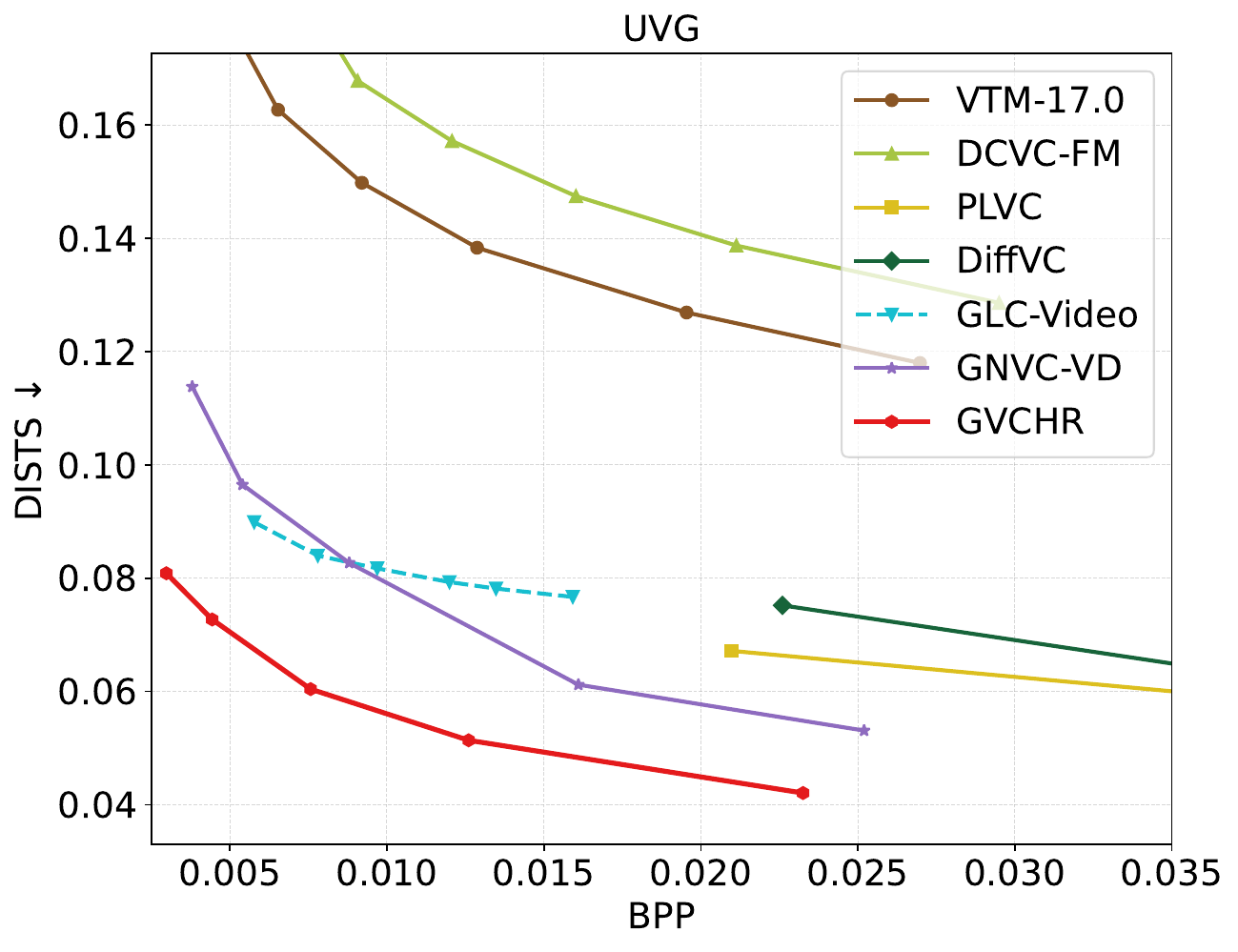}
    \label{subfig:dists_UVG}
  }
  \hspace{-10pt}
  \subfloat{
    \includegraphics[width=0.245\textwidth]{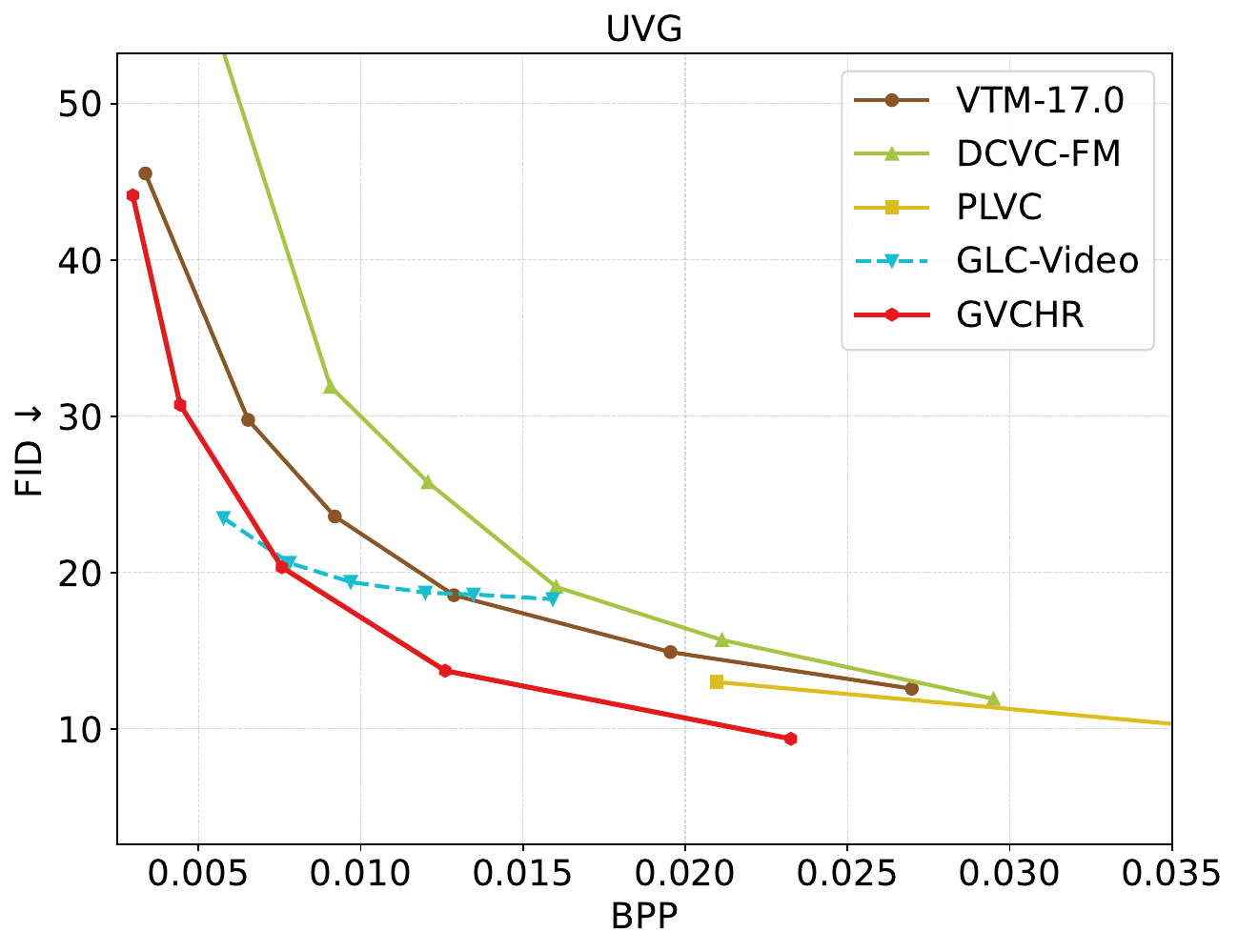}
    \label{subfig:fid_UVG}
  }
  \hspace{-10pt}
  \subfloat{
    \includegraphics[width=0.245\textwidth]{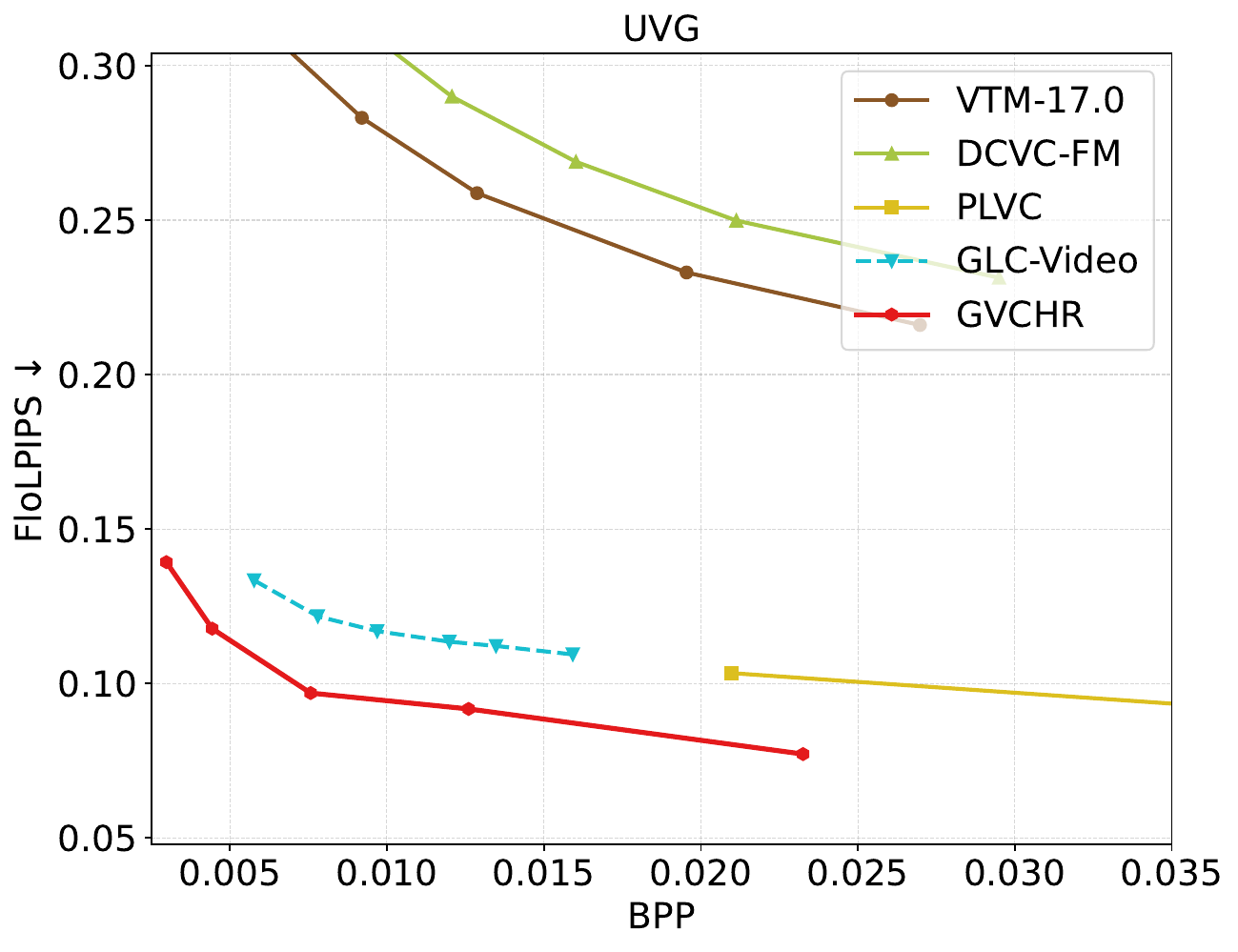}
    \label{subfig:flolpips_UVG}
  }\\[-10pt]
  
  \subfloat{
    \includegraphics[width=0.245\textwidth]{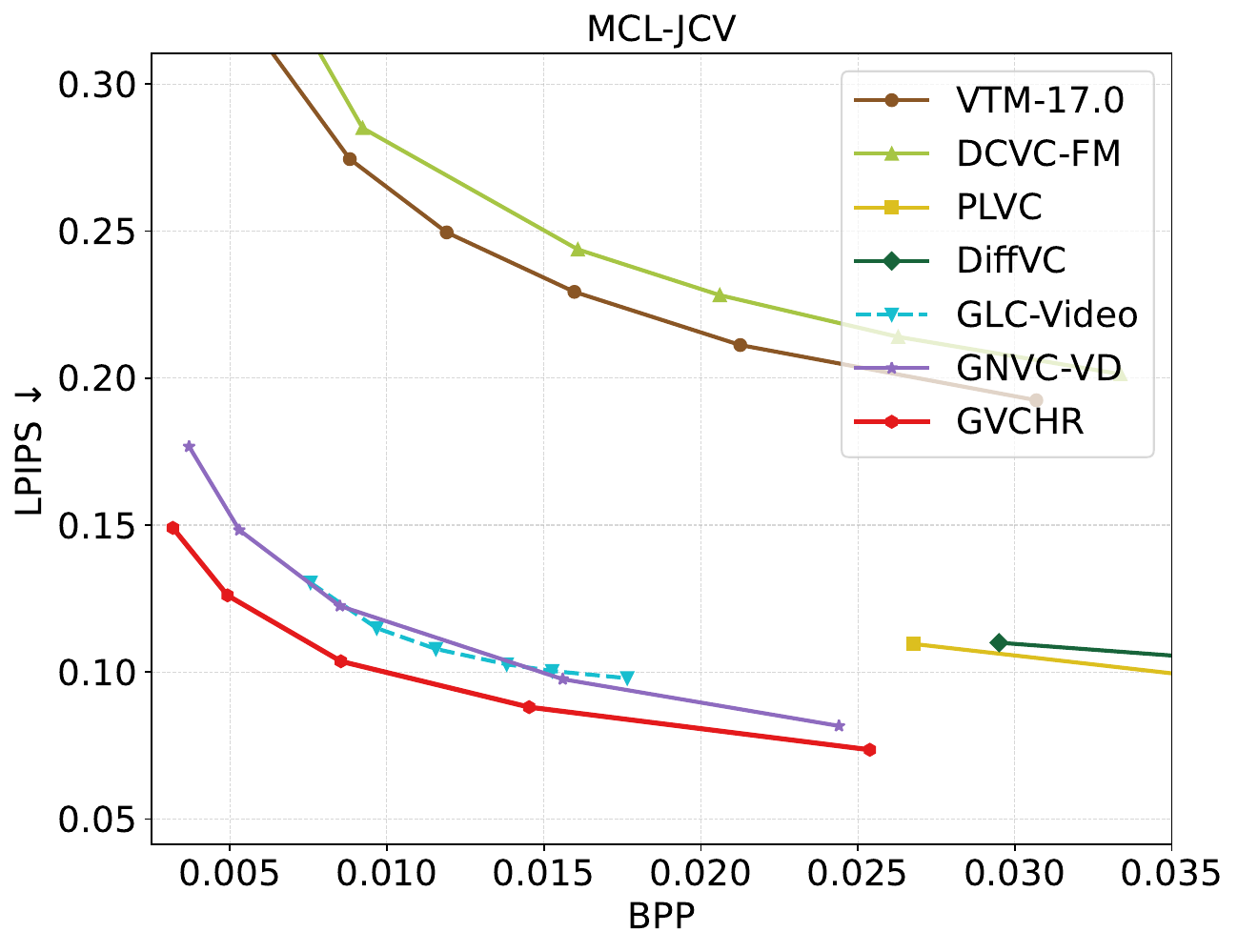}
    \label{subfig:lpips_MCL-JCV}
  }
  \hspace{-10pt}
  \subfloat{
    \includegraphics[width=0.245\textwidth]{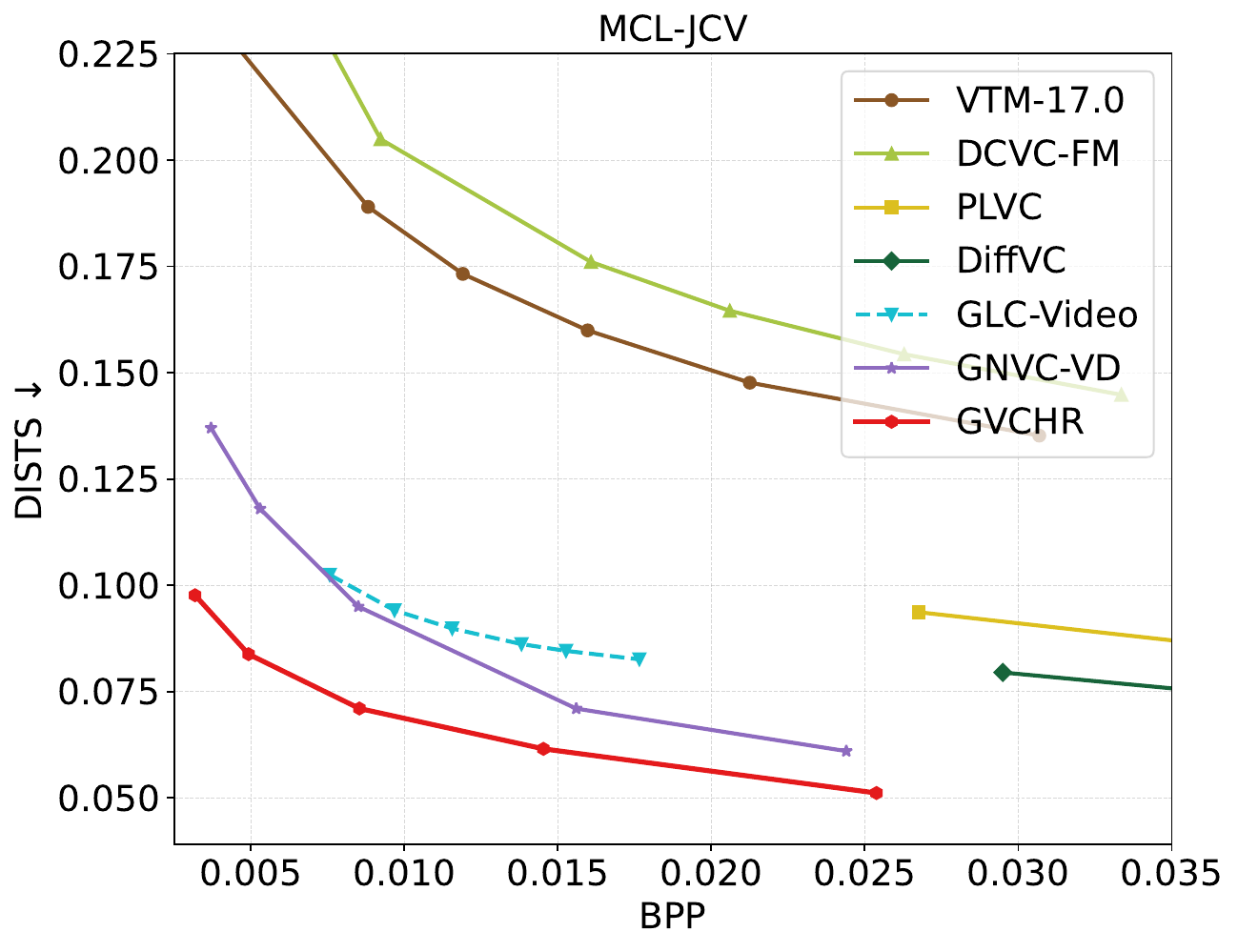}
    \label{subfig:dists_MCL-JCV}
  }
  \hspace{-10pt}
  \subfloat{
    \includegraphics[width=0.245\textwidth]{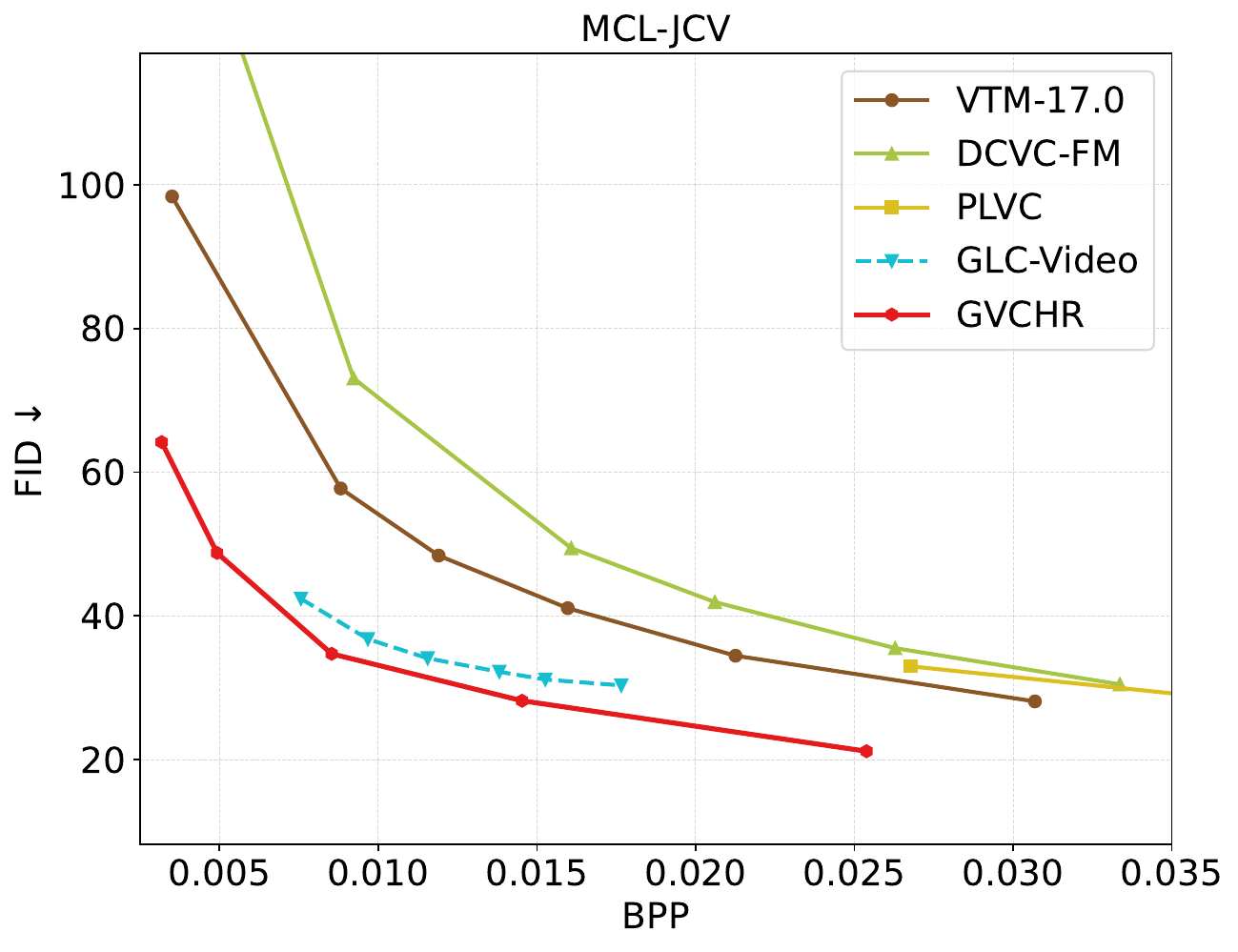}
    \label{subfig:fid_MCL-JCV}
  }
  \hspace{-10pt}
  \subfloat{
    \includegraphics[width=0.245\textwidth]{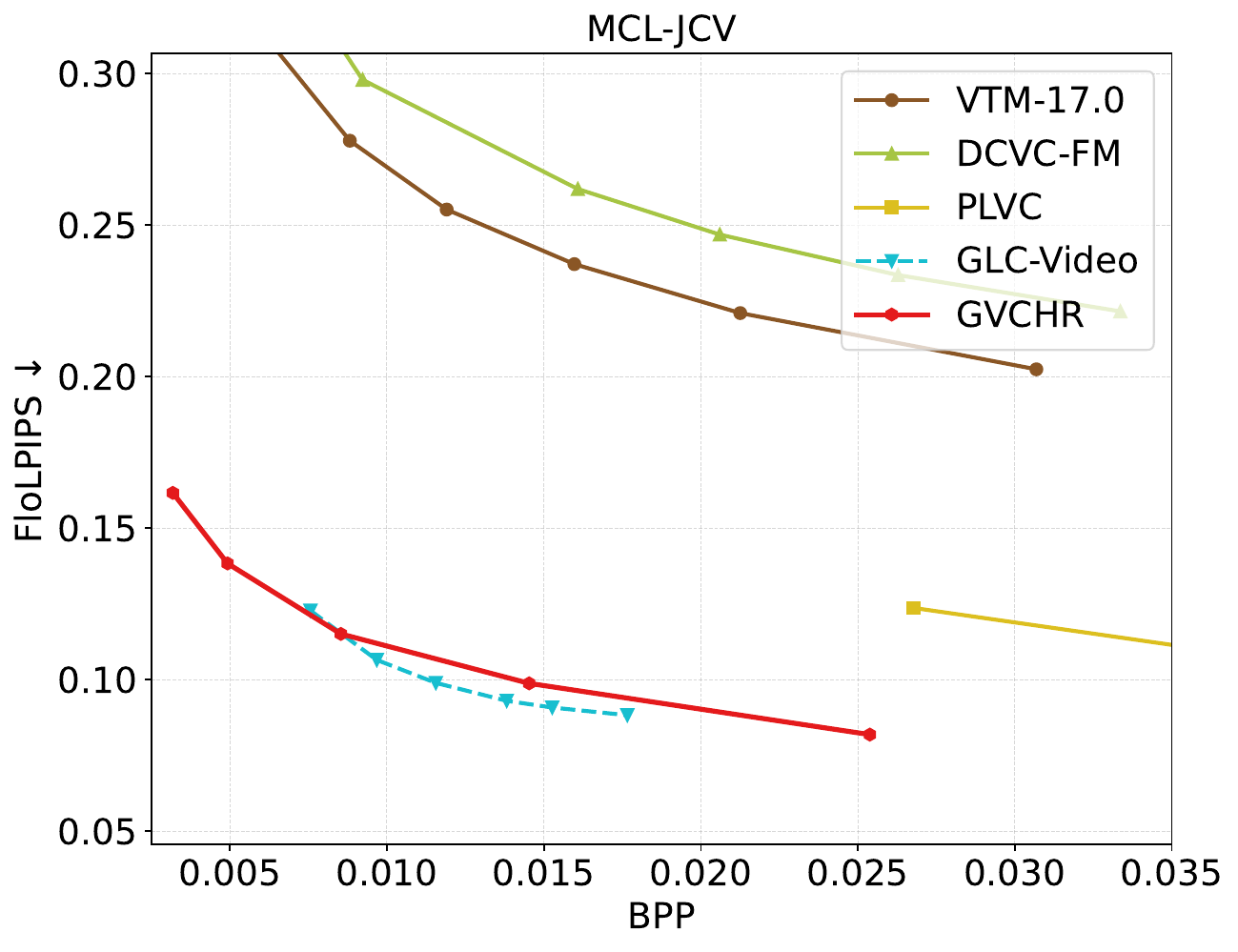}
    \label{subfig:flolpips_MCL-JCV}
  }
  \vspace{-10pt}
  \caption{Rate-perception curves on HEVC B, UVG, and MCL-JCV.}
  \label{fig:Rate-perception cuves}
  \vspace{-10pt}
\end{figure*}

To optimize \ours, we adopt a multi-stage training scheme that progressively bridges the gap between the generative prior and codec-induced distortion. Generally, the training objective is a RD loss $L_\mathrm{RD}$, defined as:
\begin{equation}
    L_\mathrm{RD} = R + \lambda D,
\end{equation}
where $R$ denotes the bitrate estimated by the latent codec. The parameter $\lambda$ controls the RD trade-off, and $D$ denotes the distortion term whose form varies across training stages. Specifically, we consider the following distortion terms:
\begin{enumerate}
    \item Latent fidelity loss $L_\mathrm{Codec}$:
\begin{equation}
    L_\mathrm{Codec} = \frac{1}{1+T/4} \sum_{t=0}^{1+T/4} w_t \cdot d_\mathrm{MSE}( \mathbf{x}_t, \hat{\mathbf{x}}_t),
\end{equation}
where $d_\mathrm{MSE}$ denotes mean square error (MSE), and $w_t$ is the hierarchical temporal weight used in the hierarchical quality structure.
    \item Conditional flow-matching loss $L_\mathrm{CFM}$:
\begin{equation}
    L_\mathrm{CFM}=\frac{\mathbb{E}_{\tau\sim\mathcal{U}(0,N)} \left\|\hat{\epsilon}(\mathbf{z}_{\tau},\tau, \mathbf{u})-\left(\mathbf{z}_N-\mathbf{z}_0\right)\right\|_2^2}{(1+T/4)\times(H/8)\times(W/8)\times 16} ,
\end{equation}
    where $\left(\mathbf{z}_N-\mathbf{z}_0\right)$ serves as the training target of velocity field, and $\hat{\epsilon}(\mathbf{z}_{\tau},\tau, \mathbf{u})$ is its corresponding estimation.
    \item Pixel-domain reconstruction loss $L_\mathrm{Rec}$: 
\begin{equation}
    L_\mathrm{Rec} = \frac{1}{1+T} \sum_{i=0}^{T} w_i^\mathrm{F} \cdot d_\mathrm{MSE}(\mathbf{F}_i, \hat{\mathbf{F}}_i),
\end{equation}
where $\mathbf{F}_i$ denotes the $i$-th frame in the source video $\mathbf{V}$, and $\hat{\mathbf{F}}_i$ is its reconstructed counterpart. We set $w_i^\mathrm{F}=w_t$ if $\mathbf{F}_i$ is encoded into $\mathbf{x}_t$.

    \item Pixel-domain perceptual loss $L_\mathrm{LPIPS}$:
\begin{equation}
        L_\mathrm{LPIPS} = \frac{1}{1+T} \sum_{i=0}^T w_i^\mathrm{F}\operatorname{LPIPS}(\mathbf{F}_i, \hat{\mathbf{F}}_i),
\end{equation}
where $\operatorname{LPIPS}$ measures perceptual quality~\cite{lpips}.
\end{enumerate}

The detailed configurations of different training stages are summarized in Table~\ref{tab:training strategy}. In the first three stages, we train only the latent codec and progressively increase the number of training frames from 5 to 33. During these stages, we resize the shorter side of each training video to 256 pixels and randomly crop $256\times256$ patches. In the fourth stage, the latent codec and HA-Adapter are jointly optimized using $L_\mathrm{Codec}$ and $L_\mathrm{CFM}$ to adapt the diffusion model to codec-distorted latents. In the fifth stage, we select five values of $\lambda$ to train distinct models corresponding to five quality levels, and further incorporate $L_\mathrm{Rec}$ and $L_\mathrm{LPIPS}$ to improve reconstruction fidelity and perceptual quality. In the sixth stage, we finetune all models at a higher resolution, remove $L_\mathrm{CFM}$, and reduce the weight of $L_\mathrm{Codec}$ so that the optimization focuses more on pixel-domain reconstruction.

\section{Experiments}\label{sec:experiments}

\begin{table*}[t]
    \begin{center}
    \caption{\label{tab:comparison}
    BD-rate ($\%$) $\downarrow$ / BD-metric $\uparrow$ on HEVC B, UVG, and MCL-JCV.
    }
    \vspace{-15pt}
    \renewcommand{\arraystretch}{1.25}
    \renewcommand{\tabcolsep}{0.5mm}
    \begin{adjustbox}{max width=\linewidth}
    \begin{tabular}{l|c|c|c|c||c|c|c|c||c|c|c|c}
    \hline
    \cline{1-13}
    & \multicolumn{4}{c||}{HEVC B} & \multicolumn{4}{c||}{UVG} & \multicolumn{4}{c}{MCL-JCV} \\
    \hline
    \cline{1-13}
    Method &
    LPIPS & DISTS & FID & FloLPIPS &
    LPIPS & DISTS & FID & FloLPIPS &
    LPIPS & DISTS & FID & FloLPIPS \\
    \cline{1-13}
    VTM-17.0
      & N/A / -0.1412 & N/A / -0.0571 & 25.6 / -3.4 & N/A / -0.1739
      & N/A / -0.1216 & N/A / -0.0667 & 37.4 / -3.4 & N/A / -0.1618
      & N/A / -0.1433 & N/A / -0.0847 & 89.9 / -15.0 & N/A / -0.1576 \\
    \hline
    DCVC-FM
      & N/A / -0.1465 & N/A / -0.0722 & 71.0 / -13.2 & N/A / -0.1862
      & N/A / -0.1311 & N/A / -0.0843 & 85.6 / -11.3 & N/A / -0.1902
      & N/A / -0.1582 & N/A / -0.1025 & 148.1 / -28.4 & N/A / -0.1823 \\
    \hline
    PLVC
      & N/A / N/A & N/A / N/A & N/A / N/A & 110.6 / N/A
      & N/A / N/A & N/A / N/A & N/A / N/A & N/A / N/A
      & 118.8 / N/A & 139.6 / N/A & 89.4 / N/A & 243.0 / N/A \\
    \hline
    DiffVC
      & - / - & - / - & - / - & - / -
      & 298.2 / N/A & N/A / N/A & - / - & - / -
      & 162.6 / N/A & N/A / N/A & - / - & - / - \\
    \hline
    GLC-Video
      & 0.0 / 0.0000 & 0.0 / 0.0000 & \textcolor{blue}{0.0} / \textcolor{blue}{0.0} & \textcolor{blue}{0.0} / \textcolor{blue}{0.0000}
      & \textcolor{blue}{0.0} / \textcolor{blue}{0.0000} & 0.0 / 0.0000 & \textcolor{blue}{0.0} / \textcolor{blue}{0.0} & \textcolor{blue}{0.0} / \textcolor{blue}{0.0000}
      & \textcolor{blue}{0.0} / \textcolor{blue}{0.0000} & 0.0 / 0.0000 & \textcolor{blue}{0.0} / \textcolor{blue}{0.0} & \textcolor{red}{0.0} / \textcolor{red}{0.0000} \\
    \hline
    GNVC-VD
      & \textcolor{blue}{-22.9} / \textcolor{blue}{0.0168} & \textcolor{blue}{-43.4} / \textcolor{blue}{0.0208} & - / - & - / -
      & 19.3 / -0.0047 & \textcolor{blue}{-2.0} / \textcolor{blue}{0.0034} & - / - & - / -
      & 0.5 / -0.0002 & \textcolor{blue}{-15.7} / \textcolor{blue}{0.0082} & - / - & - / - \\
    \hline
    \textbf{GVCHR}
      & \textcolor{red}{-63.7} / \textcolor{red}{0.0405} & \textcolor{red}{N/A} / \textcolor{red}{0.0370} & \textcolor{red}{-49.3} / \textcolor{red}{10.2} & \textcolor{red}{-30.7} / \textcolor{red}{0.0117}
      & \textcolor{red}{-65.0} / \textcolor{red}{0.0326} & \textcolor{red}{-74.0} / \textcolor{red}{0.0258} & \textcolor{red}{-7.6} / \textcolor{red}{2.6} & \textcolor{red}{-51.4} / \textcolor{red}{0.0237}
      & \textcolor{red}{-36.4} / \textcolor{red}{0.0152} & \textcolor{red}{-65.8} / \textcolor{red}{0.0251} & \textcolor{red}{-22.2} / \textcolor{red}{3.9} & \textcolor{blue}{9.7} / \textcolor{blue}{-0.0036} \\
    \hline
    \cline{1-13}
    \end{tabular}
    \end{adjustbox}
    \\[3pt]
    \parbox{\linewidth}{\footnotesize
    The best and second-best results are highlighted in \textcolor{red}{red} and \textcolor{blue}{blue}, respectively. ``N/A'' indicates that BD-rate cannot be computed due to insufficient overlap between the rate--metric curves. ``-'' denotes that the closed-source baseline does not report the result.
    }
    \end{center}
    \vspace{-15pt}
\end{table*}

\begin{figure*}[t]
    \centering
    {\footnotesize
        \begin{tabularx}{\textwidth}{
          >{\centering\arraybackslash\hsize=2.05\hsize\linewidth=\hsize}X
          >{\centering\arraybackslash\hsize=0.79\hsize\linewidth=\hsize}X
          >{\centering\arraybackslash\hsize=0.79\hsize\linewidth=\hsize}X
          >{\centering\arraybackslash\hsize=0.79\hsize\linewidth=\hsize}X
          >{\centering\arraybackslash\hsize=0.79\hsize\linewidth=\hsize}X
          >{\centering\arraybackslash\hsize=0.79\hsize\linewidth=\hsize}X
        }
        & Ground Truth & DCVC-FM & PLVC & GLC-Video & \textbf{\ours} \\
        \end{tabularx}
    }
    \vspace{-4pt}
    \includegraphics[width=\textwidth]{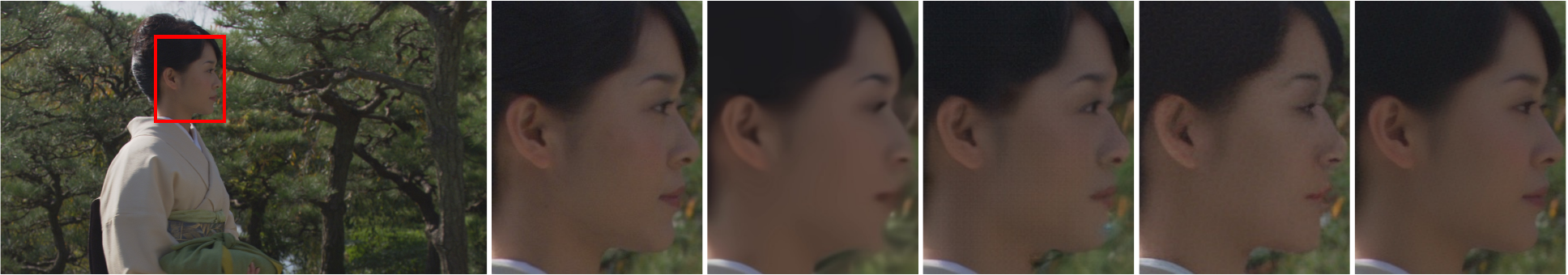}
    \vspace{2pt}
    {\footnotesize
        \begin{tabularx}{\textwidth}{
          >{\centering\arraybackslash\hsize=2.05\hsize\linewidth=\hsize}X
          >{\centering\arraybackslash\hsize=0.79\hsize\linewidth=\hsize}X
          >{\centering\arraybackslash\hsize=0.79\hsize\linewidth=\hsize}X
          >{\centering\arraybackslash\hsize=0.79\hsize\linewidth=\hsize}X
          >{\centering\arraybackslash\hsize=0.79\hsize\linewidth=\hsize}X
          >{\centering\arraybackslash\hsize=0.79\hsize\linewidth=\hsize}X
        }
        \textit{Kimono} & BPP$\downarrow$ / LPIPS$\downarrow$ & 0.0125 / 0.292 & 0.0263 / 0.155 & 0.0111 / 0.156 & \textbf{0.0103} / \textbf{0.147} \\
        \end{tabularx}
    }
    \vspace{-20pt}
    \caption{Visual comparisons with open-source baselines on the \textit{Kimono} sequence from HEVC B.} 
    \label{fig:visualization on YachtRide}
    \vspace{-10pt}
\end{figure*}

\subsection{Experimental Protocol}

\subsubsection{Training Details}
To build a high-quality training set, we curate source video clips from Pexels\footnote{\url{https://www.pexels.com}}. For data filtering, we retain clips whose aesthetic scores in VBench~\cite{vbench} are above 4.5 and clarity scores based on CLIP-IQA~\cite{clip-iqa} are above 0.65, where both metrics are averaged across all frames in each clip. We further preserve only samples with aspect ratios close to 16:9, yielding approximately 480K videos for the training.

We initialize the HA-Adapter and the video DiT with the pretrained weights from VACE~\cite{vace}, while keeping the HA-Adapter trainable and the video DiT frozen. The latent codec builds upon the architecture of DCVC-RT~\cite{dcvc-rt} and is trained from scratch. AdamW is used as the optimizer~\cite{adamw}. The latent GOP size and latent mini-GOP size are set to 8 and 4, respectively, corresponding to a GOP size of 32. Here, a GOP size of 32 denotes 32 inter-frames excluding the first intra-frame, so each coding and generation pass operates on 33 frames in total. The trade-off parameter $\lambda$ in the RD loss is sampled at five points via logarithmic interpolation over $\left[160, 3072\right]$.
Following~\cite{dcvc-rt}, each $\lambda$ is paired with a QP to control bitrate during inference. For each hierarchical temporal weight $w_t$, we assign a QP offset. Specifically, the QP offset is 8 for $w_0=8.0$. For the periodically repeated pattern $\left [ w_t, w_{t+1}, w_{t+2}, w_{t+3}\right]=\left [0.5,0.9,0.5,1.2 \right ]$ in a latent mini-GOP of size 4, the QP offsets are set to $\left [0,2,0,4 \right]$. Additional training settings are summarized in Table~\ref{tab:training strategy}.

\subsubsection{Test Details}
We evaluate \ours on the HEVC B~\cite{hevc}, UVG~\cite{uvg}, and MCL-JCV~\cite{mcl-jcv} datasets, where raw YUV420 videos are converted to RGB using BT.709~\cite{dcvc-dc}. For each sequence, the first 96 frames are tested under the GOP 32 setting. In the last segment, the final frame is replicated as padding to reach 33 frames, and the padded frames are discarded after reconstruction. We use BD-Rate and BD-Metric~\cite{bd-rate} to evaluate compression performance. Signal fidelity is measured by PSNR and MS-SSIM~\cite{ms-ssim}. Perceptual quality is assessed by LPIPS~\cite{lpips} and DISTS~\cite{dists}. To assess temporal perceptual consistency, we use FloLPIPS~\cite{flolpips}, a flow-guided LPIPS variant. To measure generative realism, we compute FID~\cite{fid} following~\cite{hific}. Bitrate is reported in bits per pixel (BPP).

\subsubsection{Baselines}
We compare \ours with three categories of baseline: 1)
the traditional video codec VTM-17.0~\cite{vvc}; 2) the fidelity-oriented NVC method DCVC-FM~\cite{dcvc-fm}; and 3) GVC methods, including PLVC~\cite{plvc}, GLC-Video~\cite{glc}, DiffVC~\cite{diffvc}, and GNVC-VD~\cite{gnvc-vd}. For VTM-17.0, we follow the evaluation protocol in~\cite{dcvc-dc}. Since the code for DiffVC and GNVC-VD is unavailable, we report the results extracted from their paper. Other baselines are tested with their default settings.

We use the latest open-source GLC-Video as the anchor for BD-rate computation, rather than VTM-17.0 as usual, because the RD curves of VTM-17.0 are often far from those of GVC methods and have limited overlap in quality range.

\subsection{Comparison Results}

\begin{figure}[t]
  \centering
  \vspace{-10pt}
  \subfloat{
    \includegraphics[width=0.235\textwidth]{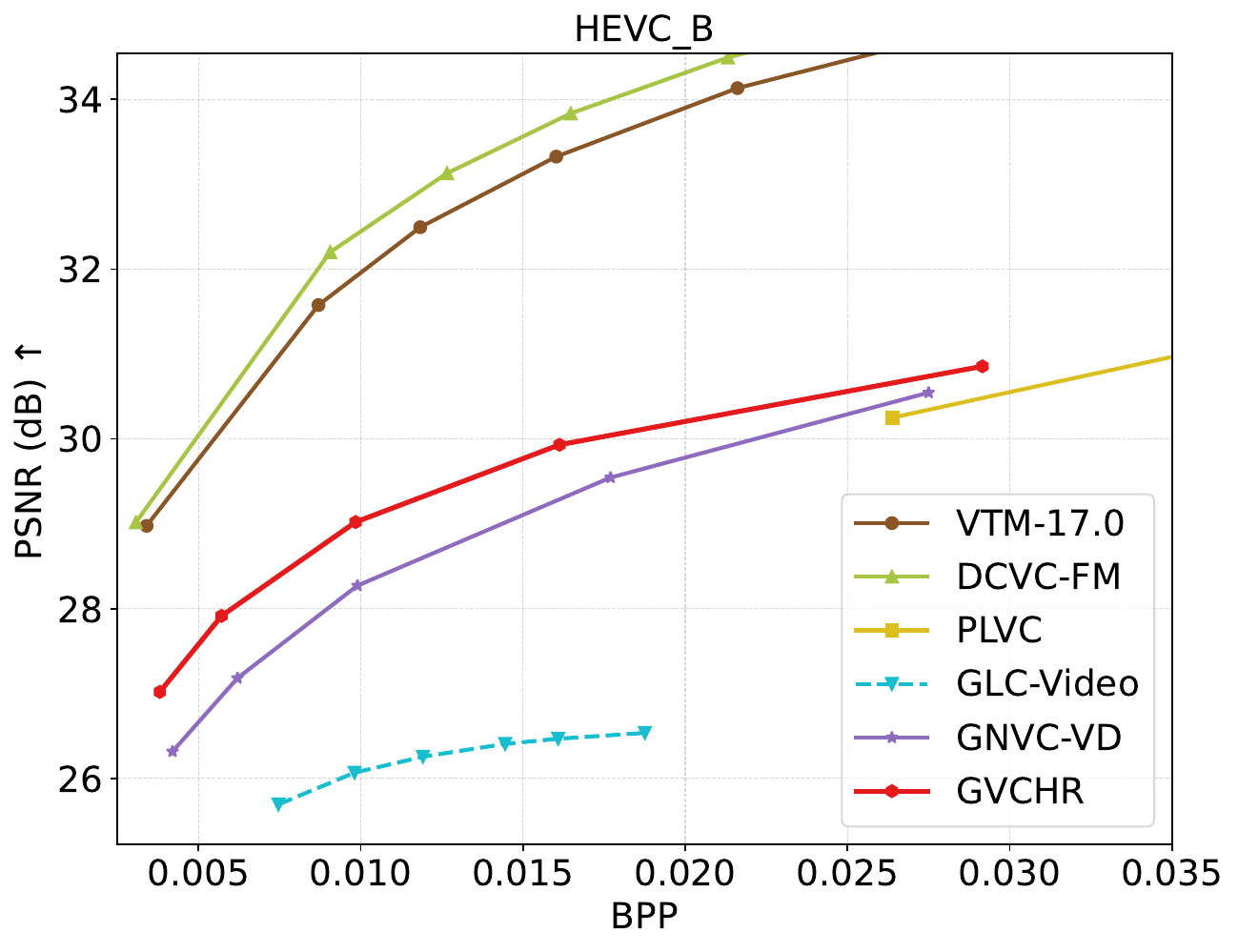}
    \label{subfig:psnr_HEVC_B}
  }
  \hspace{-10pt}
  \subfloat{
    \includegraphics[width=0.235\textwidth]{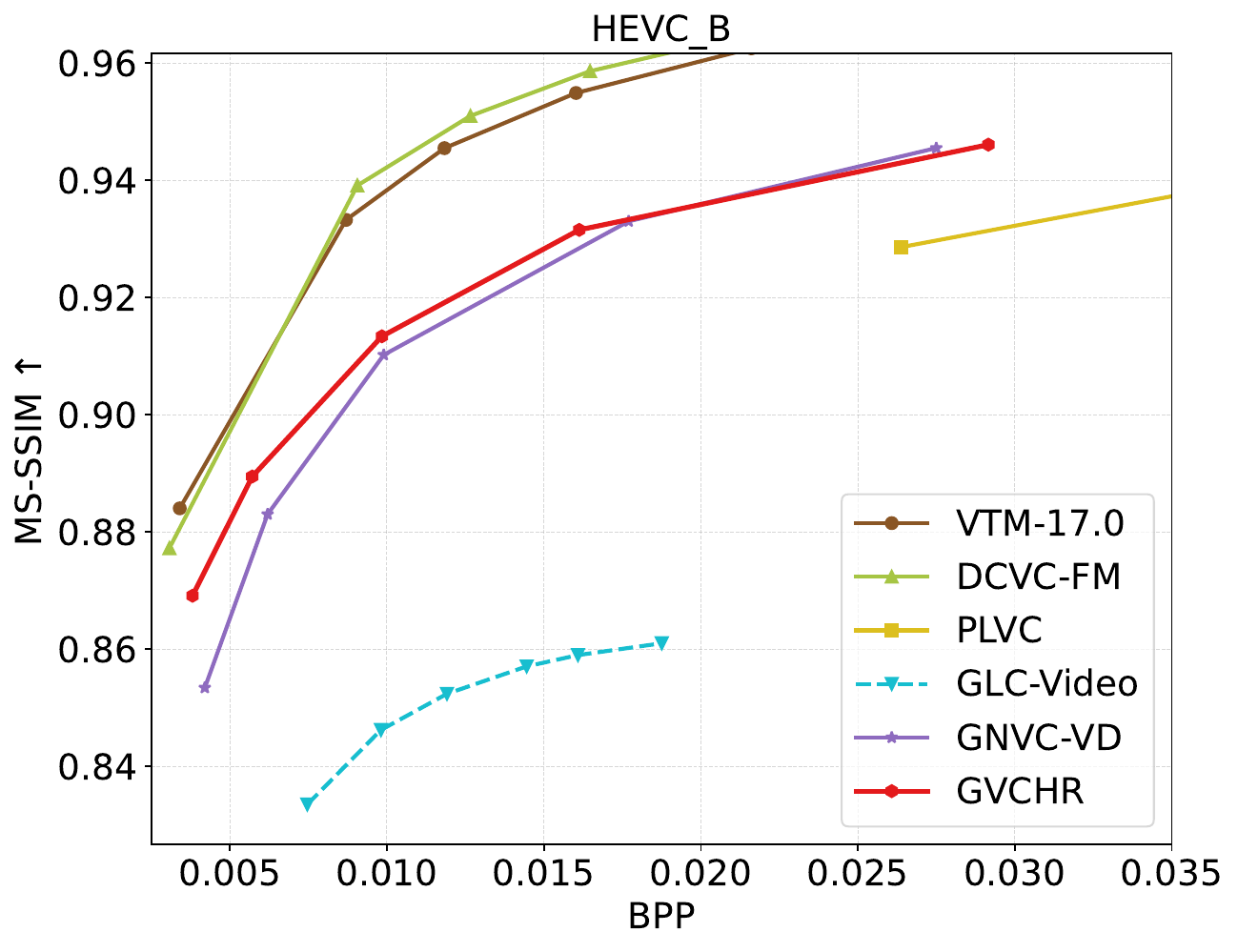}
    \label{subfig:msssim_HEVC_B}
  }\\[-10pt]
  \subfloat{
    \includegraphics[width=0.235\textwidth]{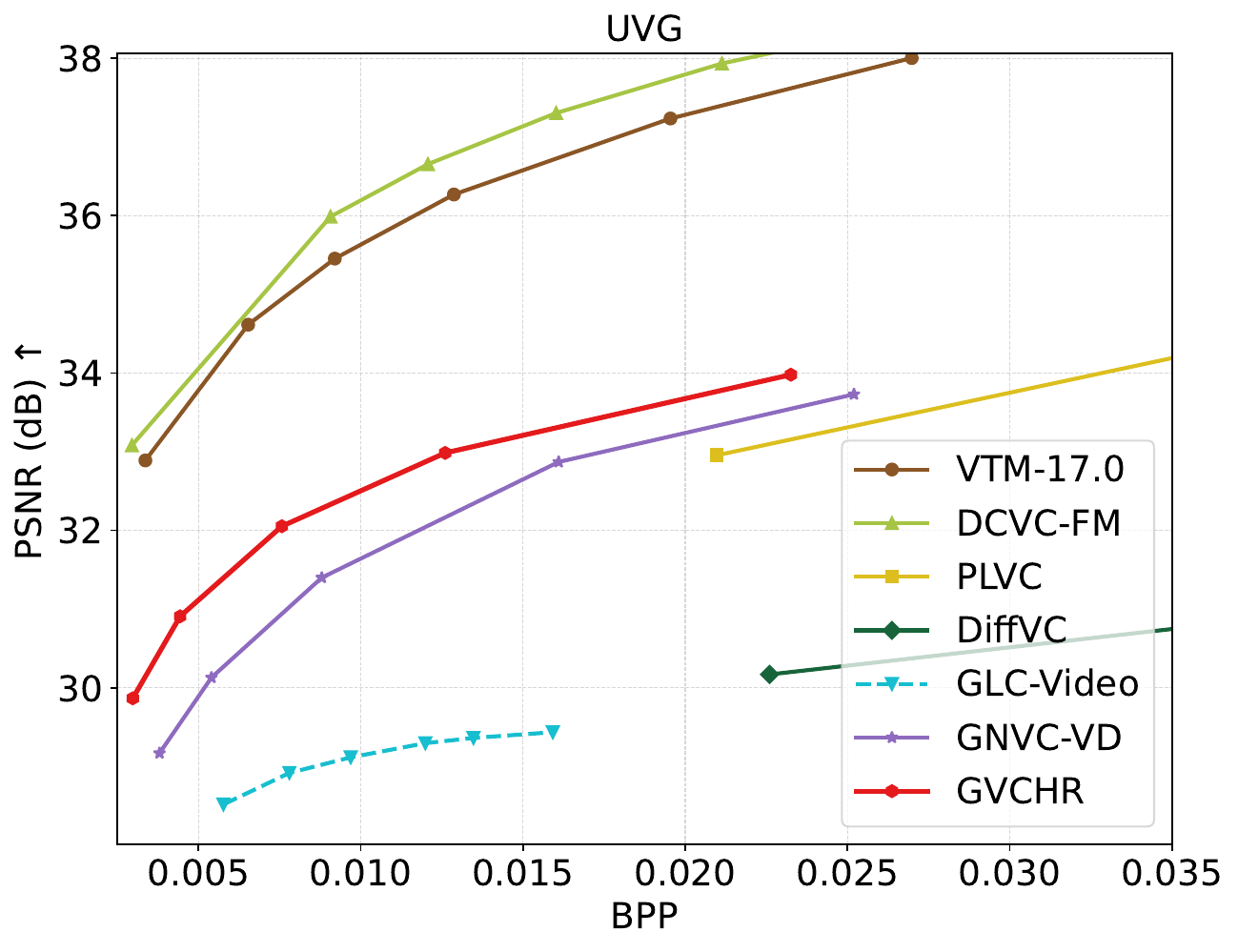}
    \label{subfig:psnr_UVG}
  }
  \hspace{-10pt}
  \subfloat{
    \includegraphics[width=0.235\textwidth]{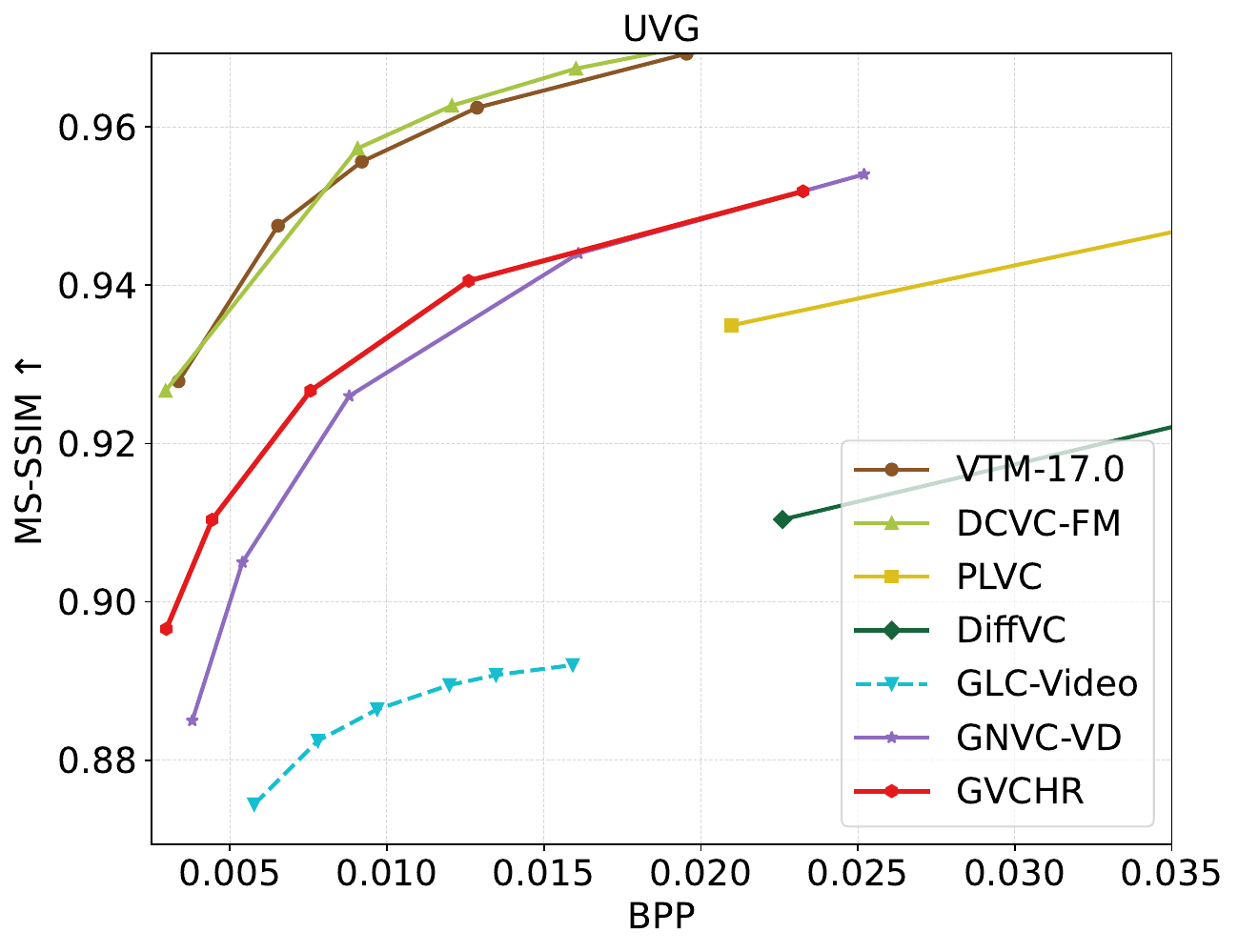}
    \label{subfig:msssim_UVG}
  }\\[-10pt]
  \subfloat{
    \includegraphics[width=0.235\textwidth]{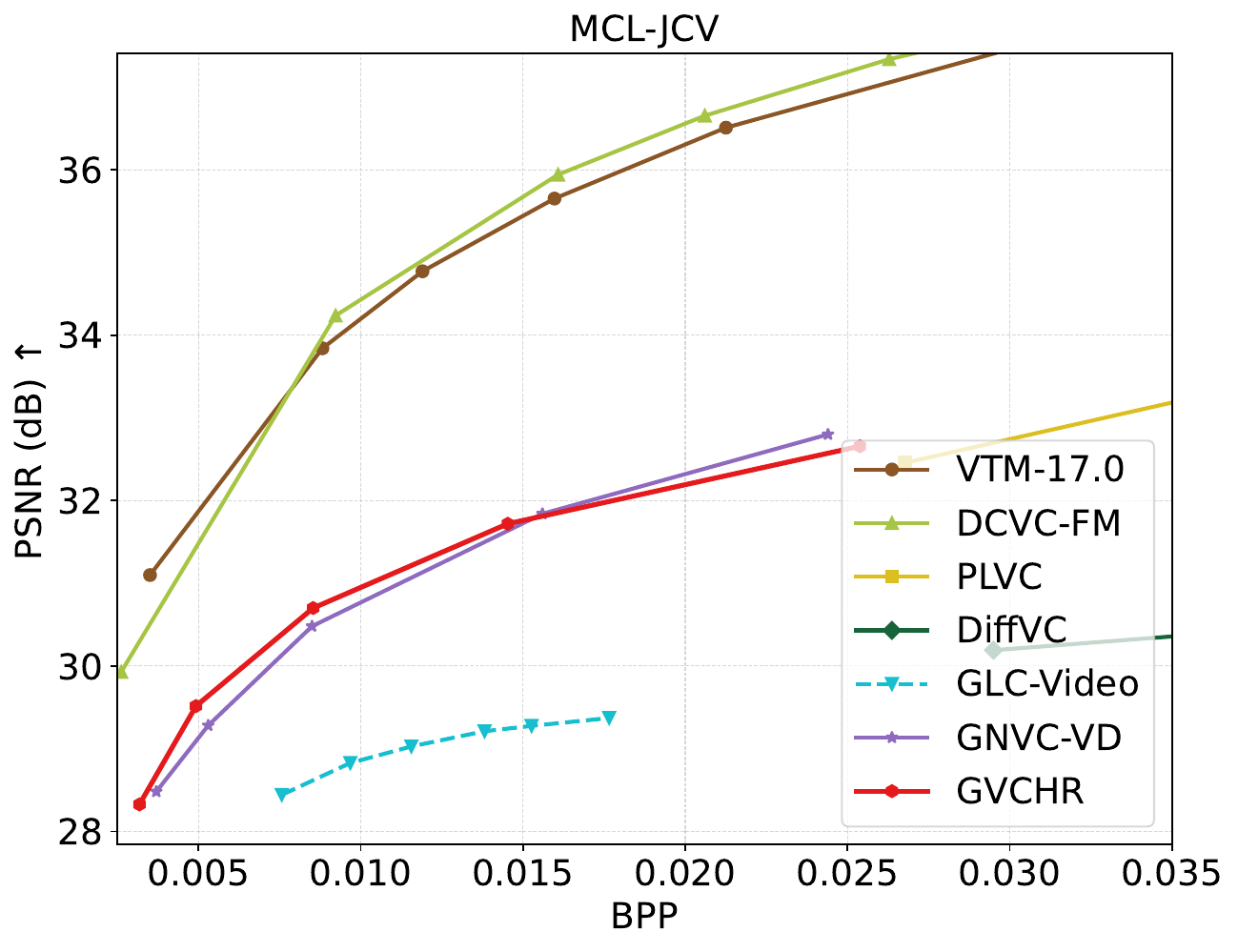}
    \label{subfig:psnr_MCL-JCV}
  }
  \hspace{-10pt}
  \subfloat{
    \includegraphics[width=0.235\textwidth]{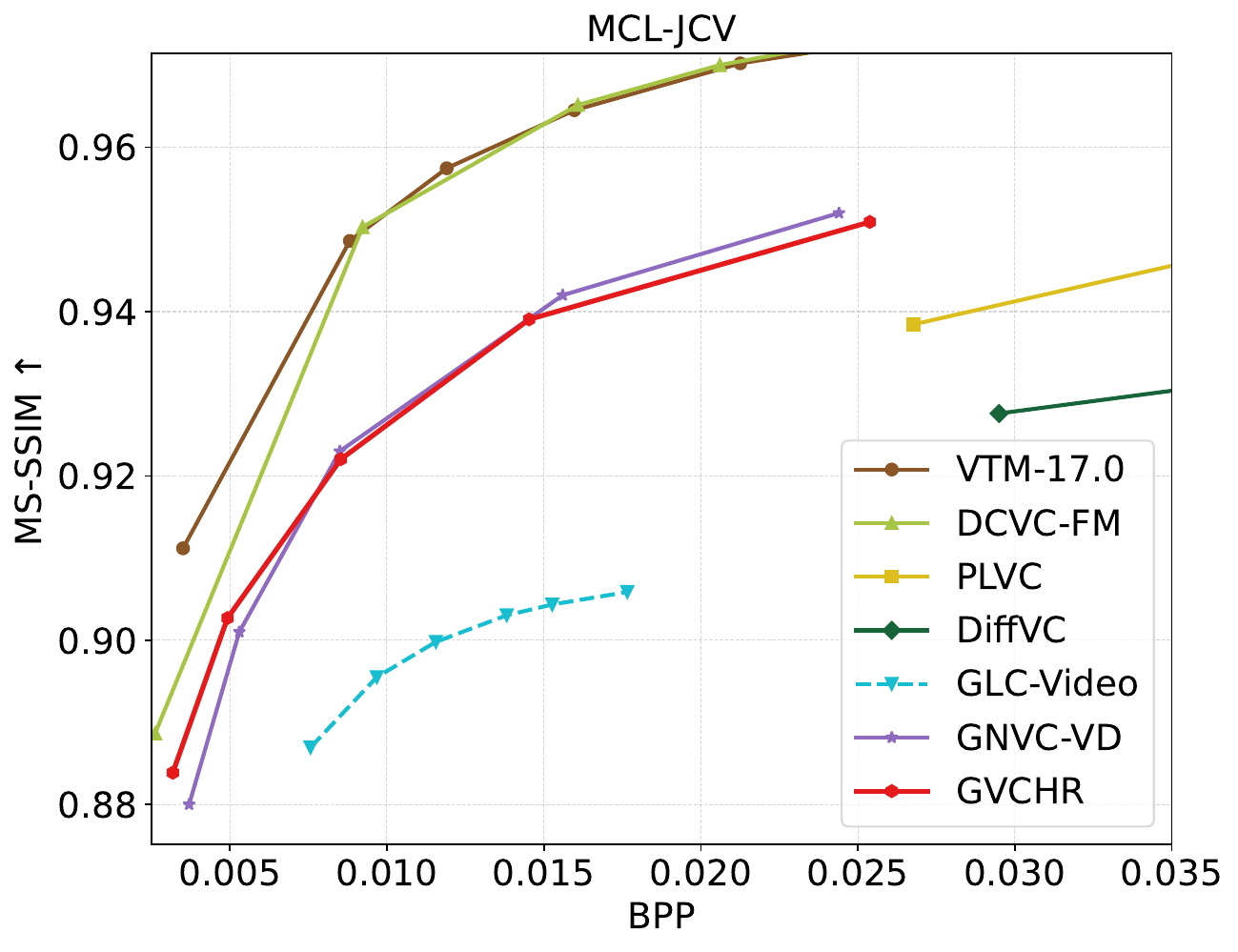}
    \label{subfig:msssim_MCL-JCV}
  }
  \vspace{-10pt}
  \caption{Rate-fidelity curves on HEVC B, UVG, and MCL-JCV.}
  \label{fig:Rate-fidelity cuves}
  \vspace{-12.5pt}
\end{figure}

\begin{figure}[t]
\centering
\includegraphics[width=\linewidth]{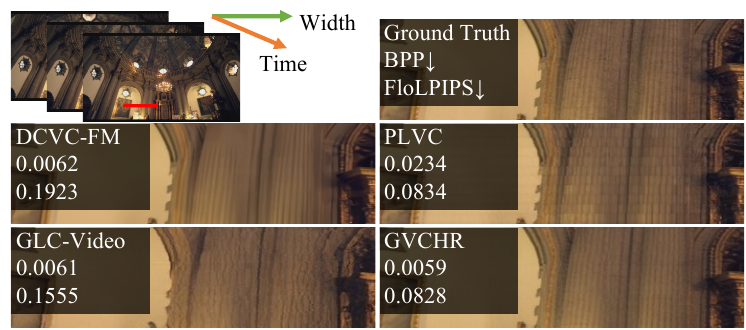}
\vspace{-20pt}
\caption{Visual comparison of temporal coherence by stacking the \textcolor{red}{red} pixel row across consecutive frames in the \textit{videoSRC15} sequence from MCL-JCV. }
\vspace{-10pt}
\label{fig:visualization of temporal coherence on YachtRide}
\end{figure}

\begin{table}[t]
\caption{BD-rate (\%)$\downarrow$ and runtime (ms) $\downarrow$ comparison with GNVC-VD.}
\centering
\vspace{-7.5pt}
\renewcommand{\arraystretch}{1.05}
\renewcommand{\tabcolsep}{3pt}
\begin{adjustbox}{max width=\linewidth}
    \begin{tabular}{c|c|c|c|c|c|c}
    \toprule
    \multirow{2}{*}{Method} & \multicolumn{4}{c|}{Averaged on three test sets}  & \multirow{2}{*}{Enc. T} & \multirow{2}{*}{Dec. T} \\
    \cmidrule{2-5}
    & LPIPS & DISTS & PSNR & MS-SSIM & & \\
    \midrule
    GNVC-VD & 0.0 & 0.0 & 0.0 & 0.0 & 439 & 4225 \\
    \textbf{\ours (GOP 24)} & -47.0 & -49.1 & -12.4 &  -4.6 & 472 & 2431 \\
    \textbf{\ours (GOP 32)} & -50.5 & -54.0 & -23.3 & -11.1 & 396 & 2814 \\
    \bottomrule
    \end{tabular}
\end{adjustbox}
\\[3pt]
\parbox{\linewidth}{\footnotesize
    ``Enc. T'' / ``Dec. T'' denotes the per-frame encoding / decoding time (ms), tested on an NVIDIA H20 GPU. We pair \textbf{GOP 24} with latent GOP 6 and latent mini-GOP 6, and \textbf{GOP 32} with latent GOP 8 and latent mini-GOP 4.
    }
\label{tab:comparison with gnvc-vd}
\vspace{-10pt}
\end{table}

\subsubsection{Quantitative Results}
As shown in Table~\ref{tab:comparison} and Fig.~\ref{fig:Rate-perception cuves}, \ours outperforms all baselines on most perceptual metrics. Compared to GLC-Video, \ours achieves BD-rate gain of 55.0\% \wrt LPIPS averaged on all three test sets. In terms of PSNR and MS-SSIM, \ours remains competitive among all GVC methods, as shown in Fig.~\ref{fig:Rate-fidelity cuves}. Additionally, we report the performance of \ours using the latest closed-source method GNVC-VD as the anchor in Table~\ref{tab:comparison with gnvc-vd}. Compared with GNVC-VD, \ours achieves average BD-rate gain of 50.5\% \wrt LPIPS under the GOP 32 setting. \ours also outperforms GNVC-VD under the GOP 24 setting, which matches the setting used in their paper. Note that GOP 24 refers to 24 inter-frames excluding the first intra-frame, and therefore corresponds to the GOP 25 setting reported by GNVC-VD which counts both intra- and inter-frames. These results demonstrate the state-of-the-art compression performance of \ours in diffusion-based GVC. 

\subsubsection{Qualitative Results}
As shown in Fig.~\ref{fig:visualization on YachtRide}, \ours reconstructs visually pleasing details with higher perceptual quality at lower bitrates. The fidelity-oriented NVC method DCVC-FM struggles to recover clear textures, resulting in blurred regions. Both PLVC and GLC-Video improve detail sharpness but introduce noticeable variegated and garbled artifacts. Furthermore, Fig.~\ref{fig:visualization of temporal coherence on YachtRide} highlights \ours's superior temporal coherence. Under cross-frame motion, DCVC-FM introduces blurring, whereas PLVC and GLC-Video exhibit flickering and temporally discontinuous textures. In contrast, \ours preserves sharp details with smoother temporal transitions.

\subsubsection{Complexity}
As shown in Table~\ref{tab:comparison with gnvc-vd}, we compare the per-frame encoding and decoding time of \ours with GNVC-VD. Since GNVC-VD is not open-source, we implement its architecture without training. Under the GOP 32 setting, \ours achieves faster encoding and decoding, owing to fewer costly intra-coding operations and the acceleration of block attention in the proposed HA-Adapter. Under the GOP 24 setting used in the GNVC-VD paper, \ours is only slightly slower in encoding. This is expected because the proposed HTCM increases encoding complexity. Meanwhile, \ours employs the intra-model of DCVC-RT, whereas GNVC-VD adopts the more complex image codec ELIC~\cite{elic}, which helps narrow the gap. Overall, \ours improves RD performance while maintaining relatively fast coding speed.

\subsection{Ablations}\label{subsec:ablation}

\begin{table}[t]
\caption{BD-rate (\%)$\downarrow$ comparison of different model variants.}
\centering
\vspace{-7.5pt}
\renewcommand{\arraystretch}{1.05}
\renewcommand{\tabcolsep}{3pt}
\begin{adjustbox}{max width=\linewidth}
    \begin{tabular}{c|c|c|c|c}
    \toprule
    \multirow{2}{*}{Method} & \multicolumn{4}{c}{MCL-JCV} \\
    \cmidrule{2-5}
    & LPIPS & DISTS & FloLPIPS & PSNR \\
    \midrule
    w/o Short-term Reference & 44.5 & 42.4 & 35.1 & 37.2 \\
    w/o Long-term Reference & 21.4  & 26.8 & 14.6 & 18.5 \\
    w/o Keyframe Sampling & 11.4 & 16.6 & 8.5 & 11.2 \\
    \textbf{Ours} & 0.0 & 0.0 & 0.0 & 0.0 \\
    \bottomrule
    \end{tabular}
\end{adjustbox}
\label{tab:ablation studies}
\vspace{-10pt}
\end{table}

\begin{table}[t]
\caption{BD-rate (\%)$\downarrow$ comparison of various configurations of hierarchical reference structures.}
\centering
\vspace{-7.5pt}
\renewcommand{\arraystretch}{1.05}
\renewcommand{\tabcolsep}{3pt}
\begin{adjustbox}{max width=\linewidth}
\begin{tabular}{c|c|c|c|c|c|c|c}
\toprule
 &\multicolumn{2}{c|}{Coding} & \multicolumn{1}{c|}{Generation} & \multicolumn{4}{c}{MCL-JCV} \\ 
 \cmidrule{2-8}
 &\makecell{Reference \\ hierarchy} & \makecell{Quality \\ hierarchy} & \makecell{Reference \\ hierarchy} & LPIPS & DISTS & FloLPIPS & PSNR\\
\midrule
(a) & \multicolumn{3}{c|}{1} & 19.3 & 20.7 & 7.1 & 17.5\\
(b) & \multicolumn{3}{c|}{2} & 9.2 & 11.8 & 5.2 & 12.4 \\
(c) & \multicolumn{3}{c|}{\textbf{3 (Ours)}} & 0.0 & 0.0 & 0.0 & 0.0 \\
(d) & \multicolumn{3}{c|}{4} & -3.1 & -2.6 & 2.4 & 6.2 \\
\cmidrule{1-8}
(e) & 1 & 3 & 1 & 12.9 & 15.2 & 9.3 & 16.4 \\
(f) & 1 & 3 & 3 & 8.4 & 9.7 & 6.8 & 10.5 \\
(g) & 3 & 3 & 1 & 7.2 & 7.4 &  5.3 & 8.1  \\
\bottomrule
\end{tabular}
\end{adjustbox}
\label{tab:configurations of hierarchical reference structures}
\vspace{-10pt}
\end{table}

\begin{figure}[t]
  \centering
  \hspace{-10.5pt}
  \subfloat{
    \includegraphics[width=0.245\textwidth]{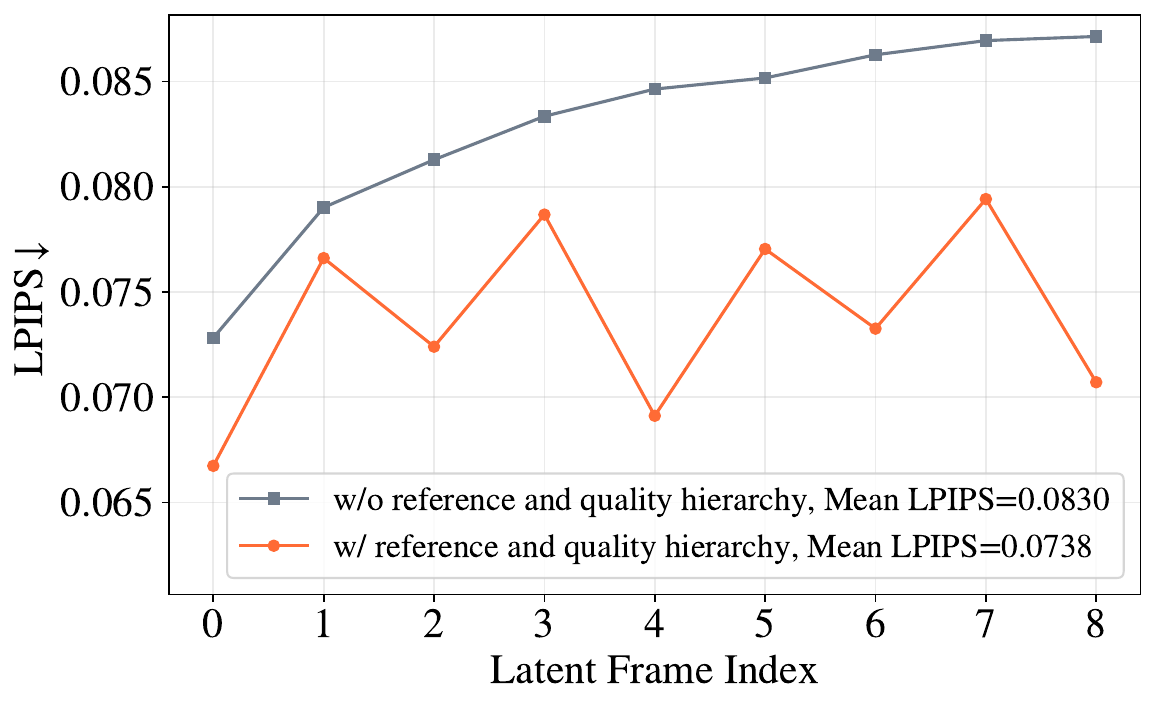}
    \label{subfig:per_frame_compare_lpips}
  }
  \hspace{-10.5pt}
  \subfloat{
    \includegraphics[width=0.245\textwidth]{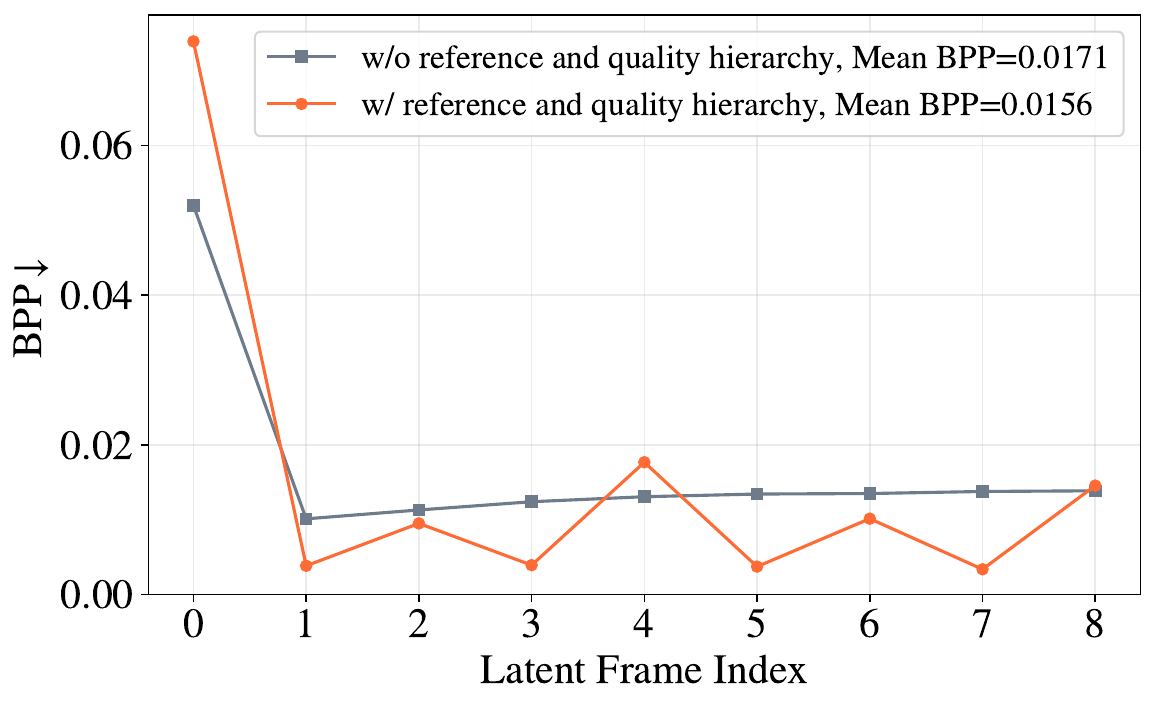}
    \label{subfig:per_frame_compare_bpp}
  }
  \vspace{-10pt}
  \caption{Perceptual quality and bit cost across latent frames within a latent GOP of the \textit{Cactus} sequence from HEVC B. Note that the LPIPS of a latent frame is averaged over its corresponding pixel-domain frames.}
  \label{fig:frame-level quality and bits}
  \vspace{-10pt}
\end{figure}

\begin{figure}[t]
\centering
\includegraphics[width=0.4\textwidth]{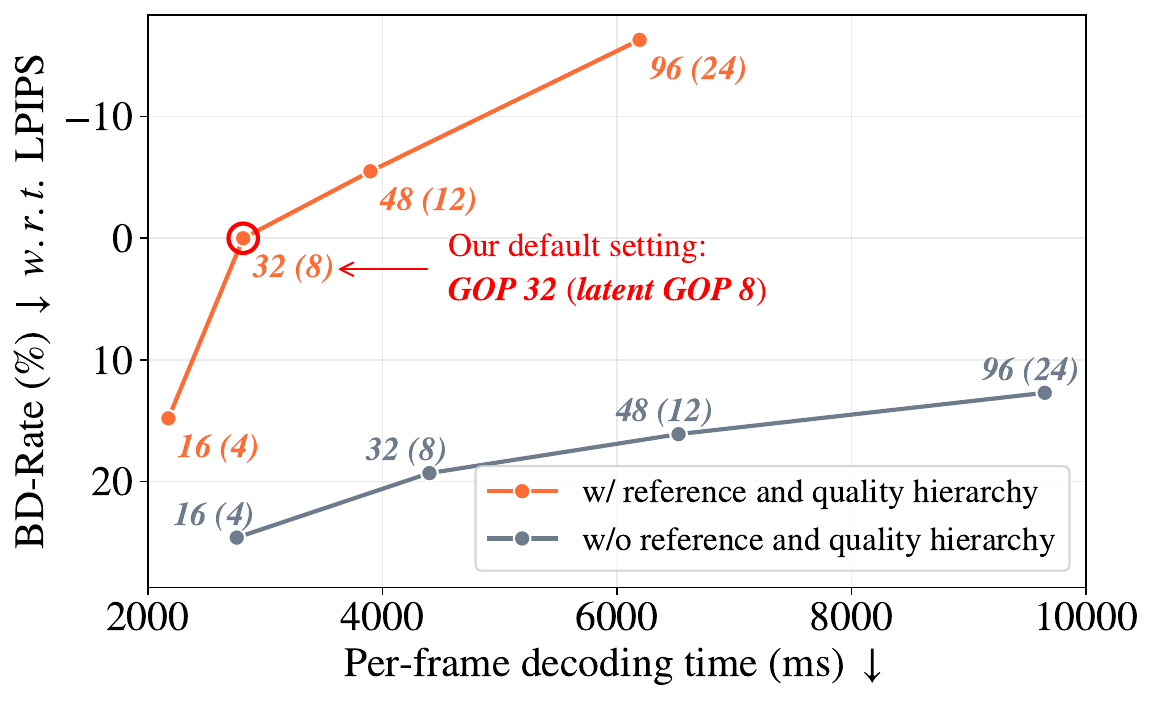}
\vspace{-10pt}
\caption{Trade-off between performance and complexity across different \textbf{\textit{GOP sizes}} (\textbf{\textit{latent GOP sizes}}) on MCL-JCV.}
\vspace{-10pt}
\label{fig:trade-off between performance and complexity}
\end{figure}

\subsubsection{Short- and Long-term References} 
We ablate short- and long-term references in HTCM. As shown in Table~\ref{tab:ablation studies}, both contribute significantly to compression performance.

\subsubsection{Hierarchical Reference Structures}\label{subsubsec: ablation on hierarchical reference structure}
As presented in Table~\ref{tab:configurations of hierarchical reference structures}, rows (a) to (d) evaluate different numbers of hierarchies in coding and generation. Notably, row (a) follows the setting of GNVC-VD, which discards both hierarchical reference and quality structures. Overall, the three-layer setting in row (c) achieves the superior performance. We further compare row (a) and row (c) in terms of frame-level perceptual quality and bit cost in Fig.~\ref{fig:frame-level quality and bits}, which shows that both reference and quality hierarchies effectively suppresses error propagation and achieves better bit allocation. Row (e) uses only the hierarchical quality structure in coding, following DiffVC. Rows (f) and (g) remove the hierarchical reference structures from coding and generation, respectively. The results in rows (e), (f), and (g) show that the hierarchical reference and quality structures in coding, together with the hierarchical reference structure in generation, are coupled and jointly contribute to compression performance.

\subsubsection{GOP Size}
Fig.~\ref{fig:trade-off between performance and complexity} presents a performance comparison of applying hierarchical reference and quality structures across different GOP sizes (or latent GOP sizes). The results demonstrate that a larger latent GOP size improves BD-Rate \wrt LPIPS, but increases decoding time, since the computational cost of attention in the generative model scales quadratically with the input GOP size. Our default setting, \ie, latent GOP 8, achieves promising compression performance while keeping decoding time relatively low. Additionally, compared with the model variant without reference and quality hierarchy, the benefits of our hierarchical designs become more pronounced as the GOP size increases. This again shows that these hierarchical structures are effective in suppressing error propagation over long dependency chains in diffusion-based GVC.

\subsubsection{Keyframe Sampling}
As shown in Table~\ref{tab:ablation studies}, removing keyframe sampling leads to inferior performance, as preserving all color information in the input increases the transmission budget. This demonstrates that \ours can recover full color information from a limited number of keyframes.

\section{Conclusion}\label{sec:conclusion}

In this paper, we propose \ours, a generative video compression framework based on hierarchical referencing. \ours organizes latent frames by reference roles and coding quality, enabling high-quality references to support efficient coding and effective generative reconstruction. In latent coding, \ours couples hierarchical reference and quality structures, and introduces HTCM to exploit complementary short- and long-term temporal context. In generative reconstruction, \ours injects decoded latents into a video DiT through a HA-Adapter, which follows the coding-side hierarchy to reduce artifact propagation during denoising. Experiments on multiple benchmarks demonstrate state-of-the-art compression performance of \ours among existing GVC methods.

\bibliographystyle{IEEEtran}
\bibliography{refs}

\end{document}